\documentclass{article} 
\usepackage[final]{colm2026_conference}

\usepackage{microtype}
\usepackage{enumitem}
\usepackage{amsmath}
\usepackage{hyperref}
\usepackage{url}
\usepackage{wrapfig}
\usepackage{multirow}
\usepackage{booktabs}
\usepackage{graphicx}
\usepackage{CJKutf8}
\usepackage[table]{xcolor}

\usepackage{lineno}

\definecolor{darkblue}{rgb}{0, 0, 0.5}
\hypersetup{colorlinks=true, citecolor=darkblue, linkcolor=darkblue, urlcolor=darkblue}

\usepackage{xcolor}

\newcommand{\revise}[1]{%
#1
}

\newcommand{\dataset}{\textsc{TIDES}}
\newcommand{\emoji}{\includegraphics[height=0.95em,trim=0 .4em 0 0]{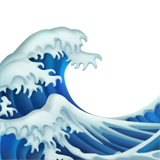}}
\newcommand{\datasetWithEmoji}{\emoji~\dataset{}}

\newcommand{\websiteicon}{%
  \raisebox{-1.5pt}{%
    \includegraphics[height=1em]{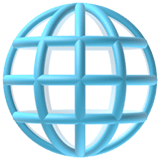}%
  }%
  \hspace{0.25em}
}

\newcommand{\githubicon}{%
  \raisebox{-1.5pt}{%
    \includegraphics[height=1.1em]{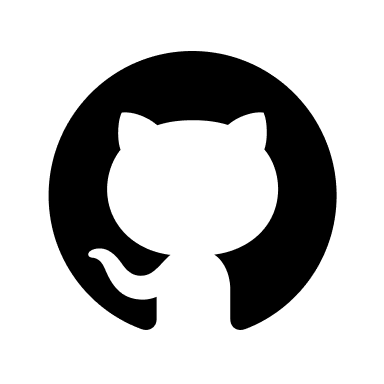}%
  }%
  \hspace{0.25em}
}

\title{\datasetWithEmoji{}: A Longitudinal Bilingual Dataset for Modeling Multi-Party Social Dynamics}

\author{Heechan Lee\textsuperscript{1,*},
     Jeonggyu Kang\textsuperscript{1,*},
     Junho Myung\textsuperscript{1}, 
     Jaywoong Jeong\textsuperscript{1},\\ 
     \textbf{Juho Kim}\textsuperscript{1,2},
     \textbf{Joseph Seering}\textsuperscript{1}\\
     KAIST\textsuperscript{1}, SkillBench\textsuperscript{2}\\
     \texttt{\{hclee99, jeonggyumarkk, seering\}@kaist.ac.kr}
}

\begin{document}

\ifcolmsubmission
\linenumbers
\fi

\maketitle
{\renewcommand{\thefootnote}{}\footnotetext{\textsuperscript{*}Equal contribution.}}

\begin{abstract}
Group conversations are fundamental to human collaboration, yet standard large language models (LLMs) still struggle with the complexities of multi-party interaction. This challenge persists in part because existing group conversation datasets are often limited to short-term lab settings with contrived tasks, failing to capture the long-term social dynamics of real-world teams. To bridge this gap, we introduce \datasetWithEmoji{}, a high-resolution longitudinal dataset tracking 12 university project teams over a full semester. Comprising \textbf{75,971} utterances in both English and Korean from in-person meetings, \dataset{} provides a naturalistic record of teams working on self-managed projects. Our socio-structural annotations---covering interaction types, emergent roles, and development stages---allow for modeling of team evolution over months. 
\revise{Experiments show that fine-tuning on \dataset{} improves next-speaker prediction by 13.8 percentage points over a bigram baseline (64.53\%) and yields performance comparable to strong proprietary zero-shot models. The model also comes within 2.1 percentage points of the published state of the art on the AMI Meeting Corpus while using approximately 42\% less training data.}
\revise{However, human evaluations suggest that better next-speaker prediction does not necessarily yield more natural or coherent utterances, as fine-tuned models were generally less preferred than vanilla models. This potential mismatch motivates further study of how structural modeling can support natural multi-party generation.}

\begin{center}
    \footnotesize
    \websiteicon
    \href{https://tides.cstlab.org}{%
        \path{tides.cstlab.org}%
    }
    \hspace{0.8em}
    \footnotesize
    \githubicon
    \href{https://github.com/cstl-kaist/TIDES_dataset}{%
        \path{github.com/cstl-kaist/TIDES_dataset}%
    }
\end{center}
\end{abstract}
\section{Introduction}

Group conversations are fundamental to human collaboration, yet standard large language models (LLMs), despite their rapid progress in dyadic settings, continue to struggle when deployed in multi-party interactions~\citep{tan-etal-2023-chatgpt, Wei2023MultiPartyCC}.
Unlike dyadic settings, successful group collaboration requires complex social coordination: understanding latent social dynamics~\citep{zhou-etal-2025-socialeval, Gu2022WhoSW} and managing turn-taking across multiple speakers~\citep{ekstedt-skantze-2020-turngpt, hilgert-niehues-2025-next, castillo-lopez-etal-2025-survey}.
This challenge is further compounded by the fact that social relationships and team dynamics are not static, but continuously evolve over prolonged interactions~\citep{Fang2025UnravelingMC}.
Developing agents capable of naturalistic multi-party interaction therefore necessitates high-quality, longitudinal resources that capture authentic social dynamics as they naturally unfold.

However, existing multi-party datasets remain insufficient for capturing the social dynamics of collaboration as they unfold in the wild.
Many widely-used datasets rely on lab-based observations~\citep{Carletta2005TheAM, Karadzhov2021DeliDataAD}, scripted media~\citep{poria-etal-2019-meld, chen-etal-2020-mpdd, zhu-etal-2021-mediasum}, or synthesized conversations~\citep{kirstein-etal-2025-need, jang-etal-2023-conversation}, and while such settings can approximate multi-party interaction, they often miss the real stakes, shared history, and evolving interpersonal dependencies that characterize authentic teamwork.
These social conditions matter because latent social dynamics---such as influence, alignment, tension, and role occupancy---do not simply appear within a single isolated exchange; they emerge through repeated collaboration as team members negotiate responsibility, expertise, and participation over time.
However, even naturalistic meeting corpora~\citep{Carletta2005TheAM, shriberg-etal-2004-icsi, Segbroeck2019DiPCoD} mostly capture only short-term episodes, limiting researchers' abilities to study how such dynamics accumulate and shift across a team's lifespan. Prior work has also centered on predefined or assigned roles~\citep{jurgens-etal-2023-spouse}, despite real teams more often exhibiting emergent functional roles that arise from ongoing interaction and changing task demands~\citep{benne1948functional}.

\begin{figure}[t]
    \centering
    \includegraphics[width=\linewidth]{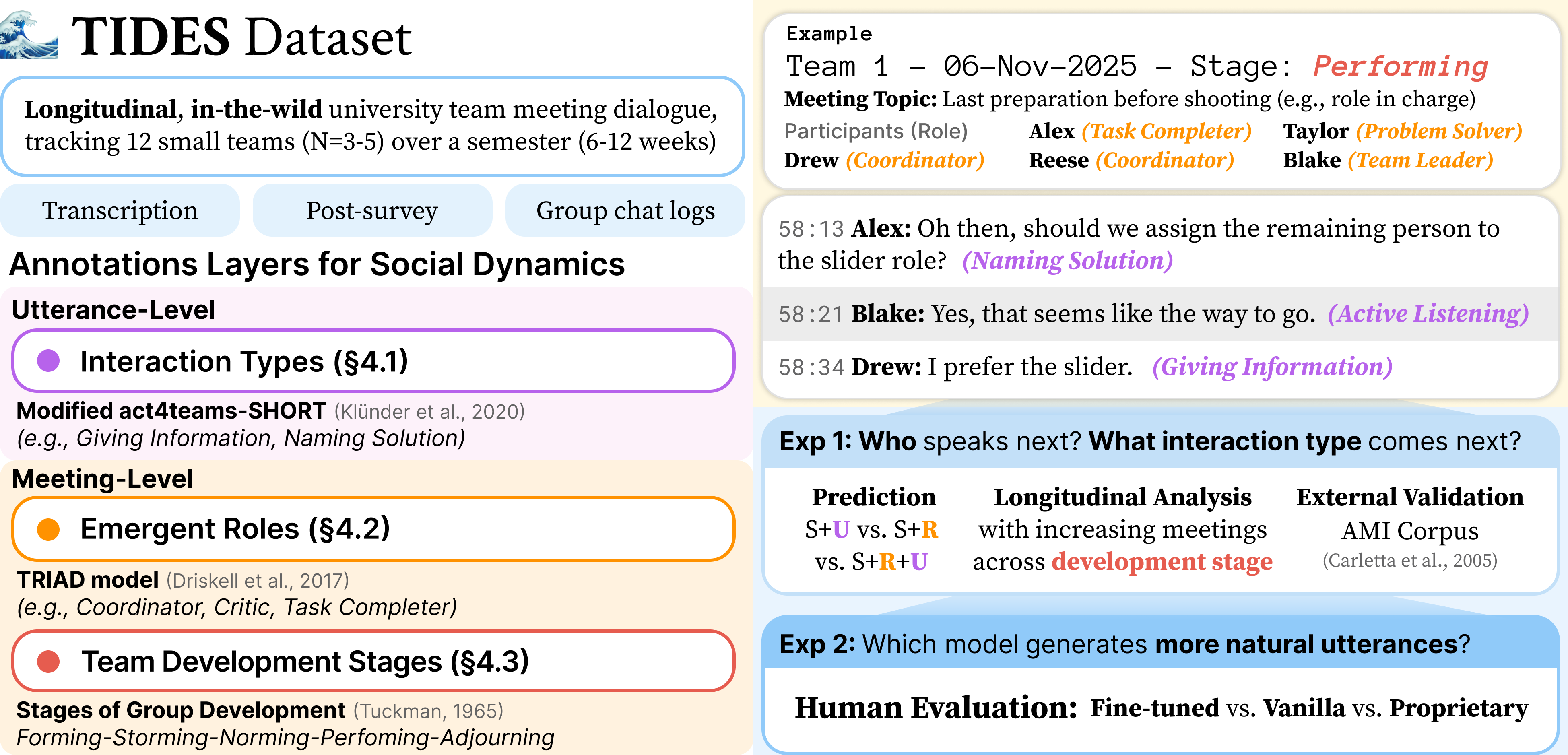}
    \caption{Overview of \dataset{}. \dataset{} is a longitudinal, in-the-wild bilingual dataset of university team project meetings, comprising 12 teams, 88 dated meetings represented in 104 transcript files, and 75,971 utterances collected over 6--12 weeks. Beyond transcripts, it provides layered annotations for social dynamics, including utterance-level interaction types, meeting-level emergent roles, and team development stages. The figure also shows an annotated example and the tasks enabled by these annotations.}
    \label{fig:main}
\end{figure}

To begin to bridge this gap, we introduce \dataset{}, \textbf{T}eam \textbf{I}nteraction and \textbf{D}ynamic \textbf{E}mergent \textbf{S}ocial-roles, a longitudinal bilingual dataset tracking real-world university project teams over a full semester. 
University courses provide a naturalistic setting where teams form organically, collaborate over weeks toward shared goals, and operate under real stakes (i.e., academic grading).
We tracked 12 teams over a semester across a diverse range of university classes, recording their actual project meetings to collect a total of \textbf{75,971} utterances from 88 dated meetings (about 110 hours), \revise{represented in 104 transcript files because some longer meetings were split into multiple parts.}
To capture not only what teams said but also how their latent social dynamics evolved, participants completed post-meeting surveys after each session, reporting collaboration satisfaction and their perceptions of each member's emergent role.
The resulting dataset pairs privacy-preserved transcripts with longitudinal social annotations, including meeting-level emergent roles, utterance-level interaction types, team development stages, and silence gaps (annotated examples in Appendix~\ref{appendix:transcript-examples}).

The longitudinal nature of \dataset{}, in combination with the social annotations, enables us to take steps toward a vision of more socially-aware LLMs for group conversation, addressing four initial questions:
(1)~\textit{Can models learn turn-taking and interaction patterns from longitudinal team data?} We evaluate next-speaker and next-intention prediction under three context conditions that vary the balance between utterance content and social structure.
(2)~\textit{How quickly does performance adapt to a specific team as more meetings are observed?} We train on incrementally more meetings from individual teams, measuring when prediction performance plateaus.
(3)~\textit{Do these patterns generalize to unseen meeting corpora?} We evaluate on the AMI Meeting Corpus, directly comparing against published state-of-the-art performance.
(4)~\textit{Does structural understanding translate to generating utterances that humans find natural?} We compare fine-tuned, vanilla, and proprietary models through automatic metrics and a human evaluation on Prolific.

\revise{Experiments show that models fine-tuned on \dataset{} improve prediction of team-level turn-taking patterns.
For next-speaker prediction, the primary fine-tuned condition reaches 64.53\%, a 13.8 percentage-point improvement over the bigram baseline and performance comparable to strong proprietary zero-shot models.
In chronological single-team analyses, most observed gains occur within the first three to four meetings, which we treat as descriptive evidence of rapid team-specific adaptation (\S\ref{result:1c}).
Through transfer to the AMI Corpus~\citep{Carletta2005TheAM}, our model achieves comparable performance to the published state of the art while using approximately 42\% less training data.}
\revise{However, preference ratings from human evaluators suggest that improvements in conversational structure prediction do not necessarily lead to more human-preferred utterance generation.
Although the fine-tuned models outperform the baselines in predicting the correct next speaker, human judges tend to prefer the vanilla models in terms of naturalness and coherence across multiple dimensions.
These findings indicate a potential mismatch between optimizing for conversational structure and producing content that users perceive as natural, motivating further investigation into how LLM agents can generate more natural utterances and what kinds of naturalness users expect from them.}

While our experiments primarily evaluate local prediction and generation within short context windows, the longitudinal annotations in \dataset{} provide the foundation for modeling how team dynamics evolve across meetings; initial evidence, such as the correlation between development stage and prediction accuracy (Appendix~\ref{appendix:detailed-breakdowns}), begins to support this direction.

\section{Related Work}

\paragraph{Multi-Party Dialogue Datasets.}
Prior work has introduced a range of multi-party meeting and discussion datasets, including the AMI and ICSI Meeting Corpora~\citep{Carletta2005TheAM, shriberg-etal-2004-icsi}, which provide foundational recordings of real meetings.
More recent resources such as MeetingBank~\citep{hu-etal-2023-meetingbank}, QMSum~\citep{zhong-etal-2021-qmsum}, and ELITR~\citep{nedoluzhko-etal-2022-elitr} have expanded the scale and utility of meeting corpora, particularly for summarization and automatic minuting, while datasets like DeliData~\citep{Karadzhov2021DeliDataAD} focus on deliberative problem-solving interactions and MPDD~\citep{chen-etal-2020-mpdd} and FAME~\citep{kirstein-etal-2025-need} provide controlled or synthetic settings for analyzing interpersonal dynamics. 
However, most existing resources are limited in at least one critical dimension: most capture short-term interactions rather than longitudinal collaboration, some are drawn from scripted or institutional settings that do not fully reflect authentic teamwork, and others rely on assigned, fictional, or synthetic roles rather than emergent team dynamics. 
In contrast, \dataset{} captures real-world collaborative behavior \textit{in the wild}, at scale and over a longer time span, enabling the study of how social roles, interaction patterns, and team development evolve throughout the teamwork.

\paragraph{Socially-Aware Dialogue Modeling.}
Understanding group dynamics computationally requires more than just dialogue transcripts. It demands annotations of the social structures that shape interaction.
Organizational science has long studied how teams develop over time, from Tuckman's stage model~\citep{Tuckman1965DEVELOPMENTALSI} to emergent role theories~\citep{kauffeld201821}, yet these frameworks have seen limited adoption in NLP, due to the lack of suitably annotated data.
Existing benchmarks for social intelligence rely on dyadic or scripted conversations~\citep{Zhou2023SOTOPIAIE, Zhan2023SocialDialAB, rashid-blanco-2018-characterizing, tigunova-etal-2021-pride, Jia2020DDRelAN, jurgens-etal-2023-spouse}, which do not capture the longitudinal, multi-party dynamics that characterize real teamwork.
\dataset{} addresses this with socio-structural annotations---interaction types, emergent roles, and development stages.
These annotations enable models to tackle core aspects of group dynamics, such as predicting who speaks next and what type of contribution they make.

\paragraph{Conversational Agents in Group Settings.}
From an HCI perspective, researchers have investigated how conversational agents can support cooperation and facilitate participation in group discussions~\citep{Claggett2025RelationalAF, Kim2020BotIT, Houde2025ControllingAA}, with a complementary line equipping agents with internal reasoning about \textit{when and why} to speak~\citep{Liu2025InnerThoughts}. 
Such approaches assume that richer reasoning about group structure will lead to more natural contributions.
\revise{However, systematic human evaluation of this assumption in multi-party settings remains limited, as prior studies often rely on structural metrics or LLM-based judges and are typically conducted in simulated or short-term settings.
By combining longitudinal social annotations with human evaluation, \dataset{} enables the relationship between structural understanding and natural utterance generation to be examined directly.}

\section{Data Collection}
\revise{We collected authentic, longitudinal team dynamics by tracking real university project teams throughout a semester. Teams formed organically, determined project goals and milestones within the guidelines of different course projects, collaborated under real stakes (i.e., grade evaluation), and interacted repeatedly over weeks toward shared goals.}

\subsection{Collection Procedure}
\revise{We recruited 12 student teams (total N = 50 recruited participants) enrolled in university courses at full-time Korean universities during the Fall 2025 semester. Participants were recruited through university- and student-managed announcement boards and lists.}
Teams ranged from 3 to 5 members, and their projects ranged in duration from 6 to 12 weeks across a variety of disciplines, including Design, Computer Science, and Industrial Engineering.
Prior to the study, all participants attended an orientation session where they were briefed on data collection procedures.
Throughout the semester, teams were asked to keep audio recordings of all team project meetings and to submit the recordings to the research team after each session.
Following each meeting, every team member individually completed a post-meeting survey so that we could capture meeting-specific perceptions of collaboration and track how team dynamics changed over time.
At the end of the semester, each team completed a final team-level survey to reflect on their overall collaboration and project outcome after their last meeting.
Additionally, chat message logs regarding the project between team members were collected from all but one team, which did not use a messaging platform during the project.
This experiment was approved by our institution's Institutional Review Board and we compensated each team with 500,000 KRW (approximately 325 USD).

\subsection{Post-processing}
\label{sec:postprocessing}

Converting 110+ hours of bilingual group conversation audio into research-ready transcripts required a six-stage pipeline.
We first transcribed all recordings with Whisper Large-V3~\citep{radford2023robust}, then assigned speaker labels through a custom diarization pipeline combining pyannote 3.1~\citep{bredin2023pyannote,plaquet2023powerset} VAD with ECAPA-TDNN~\citep{desplanques2020ecapa}, re-clustering to the known team size.

Next, we removed personally identifiable information using Microsoft Presidio~\citep{presidio} for the five English-speaking teams and a locally-run Qwen3-30B-A3B~\citep{qwen3} for the seven Korean-speaking teams, where no comparable off-the-shelf tool exists.
All Korean transcripts were then translated to English by the same local LLM.
To resolve the cross-session speaker identity problem---where the same person may receive different labels across recordings---we unified speaker labels using WeSpeaker~\citep{wang2022wespeaker} embeddings and the Hungarian algorithm~\citep{kuhn1955hungarian}, producing 422 consistent mappings across 12 teams.
\revise{Then, at least one representative from each team validated the full transcript against the original audio recordings and corrected errors. Validators were compensated at 35,000 KRW (approximately 23 USD) per hour of reviewed audio.}
\revise{The authors manually verified all data to prevent leakage of personal information, and redacted eight utterances containing offensive speech.}
\revise{After transcript anonymization, we refined utterance boundaries with GPT-5-mini guided by Language Development Project transcription rules~\citep{macwhinney2000childes}, yielding 1,238 splits and 309 merges across the processed corpus.}

Full pipeline details, model configurations, and processing statistics are provided in Appendix~\ref{appendix:postprocessing}.
\section{Data Annotation}
\label{sec:annotation}
After collecting and post-processing the recordings, we constructed multiple annotation layers to capture social dynamics in small teams.
These annotations drew on different sources: meeting transcripts for utterance-level interaction types, post-surveys for emergent role assignment, and end-of-semester group reflection for meeting-level team development stages.
\revise{As these layers capture complementary constructs from different perspectives and timescales, they should be interpreted according to their sources rather than as interchangeable measurements of a single latent social state.}

\subsection{Utterance-level Interaction Type}

\revise{We annotated utterance-level interaction types using a modified version of \texttt{act4teams-SHORT}~\citep{klunder2020you}. 
Derived from the original \texttt{act4teams} taxonomy~\citep{kauffeld201821}, this taxonomy was designed to capture interaction behaviors in team problem-solving and collaboration.
Although we considered standard dialogue-act taxonomies such as AMI-DA~\citep{hain20072007} and MRDA~\citep{shriberg-etal-2004-icsi}, we selected an act4teams-based scheme because our goal was to characterize collaboration-oriented intentions rather than general conversational or discourse functions.}
However, because \texttt{act4teams-SHORT} largely omits the social activity dimension, we extended it to better capture the social signals.
\revise{Specifically, drawing from the original \texttt{act4teams} taxonomy~\citep{kauffeld201821}, we separated \textit{Counterproductivity} into \textit{Social Negative} and \textit{Task/Process Negative}, and readopted \textit{Social/Humor}, \textit{Active Listening}, and \textit{Other/Neutral}.}
This resulted in our modified act4teams-SHORT scheme with a total of 15 categories.

To construct human-labeled gold data, we recruited 260 annotators through Prolific and assigned three independent annotators to each utterance.
We then aggregated labels using majority voting and retained labels agreed upon by at least two of the three annotators, resulting in 5,705 human-labeled gold utterances ($\simeq$ 7.5\% of all utterances).
We then fine-tuned \texttt{Gemma-3-12B} on the gold data and evaluated several candidate models under a leave-one-team-out setup, in which one team was held out from fine-tuning to reduce data leakage across teams.
We selected \texttt{Gemma-3-12B} for large-scale annotation because it achieved the highest macro-F1 score (0.54).
\revise{In addition, two authors jointly reviewed sample outputs from the candidate models to confirm the plausibility of the resulting annotations before large-scale application. Considering the complexity of the 15-category taxonomy and the subjective nature of the task, and the moderate agreement among human annotators (Fleiss' $\kappa$ = 0.400)~\citep{wong-etal-2021-cross}, we used the fine-tuned model to annotate the remaining 70K+ utterances.}
Additional details on the crowdsourcing process, model selection, and analysis on gold data are provided in Appendix~\ref{appendix:utterance-level}.

\subsection{Emergent Roles}
We assigned one peer-perception-based emergent role to each participant at each meeting.
In the survey, participants evaluated their teammates along three dimensions in the TRIAD model~\citep{Driskell2017TeamRA}---\textit{Dominance, Sociability, and Task Orientation}---using nine 5-point Likert items (three items per dimension).
We chose this dimension-based design rather than asking participants to directly assign one of the 13 TRIAD roles, as direct role categorization is difficult for non-expert participants.
Detailed survey items are provided in Appendix~\ref{appendix:full-survey}.

Because our survey used a 5-point Likert scale whereas TRIAD roles are defined in a different coordinate space (7-point Likert scale), we normalized the survey scores before role assignment.
We used team-specific cumulative normalization, where each participant's score at meeting \(m\) was standardized using the responses from the same team observed up to and including that meeting:
\begin{equation}
z^{(k)}_{t,m,i}=\frac{x^{(k)}_{t,m,i}-\mu^{(k)}_{t,\le m}}{\sigma^{(k)}_{t,\le m}}.
\end{equation}
Here, \(x^{(k)}_{t,m,i}\) denotes participant \(i\)'s average survey score on dimension \(k\) in team \(t\) at meeting \(m\), and \(\mu^{(k)}_{t,\le m}\) and \(\sigma^{(k)}_{t,\le m}\) denote the mean and standard deviation of that dimension computed from all responses in team \(t\) up to meeting \(m\).
We then assigned each participant the role whose TRIAD prototype, $\tilde{\mathbf{p}}_r$, was closest in Euclidean distance to the normalized score vector with both represented in the same standardized coordinate space:
\begin{equation}
\hat{r}_{t,m,i}=\arg\min_{r \in \mathcal{R}} \left\| \mathbf{z}_{t,m,i}-\tilde{\mathbf{p}}_r \right\|_2.
\end{equation}
\(\hat{r}_{t,m,i}\) denotes the emergent role assigned to participant \(i\) in team \(t\) at meeting \(m\).
To support alternative analyses beyond discrete role labels, \dataset{} also includes the raw scores on the three dimensions for each participant at each meeting.

\subsection{Team Development Stage Annotation}

To annotate how teams evolved over time, we assigned each meeting a stage from Tuckman's team development model: Forming, Storming, Norming, Performing, or Adjourning~\citep{Tuckman1965DEVELOPMENTALSI}.
During the final post-data collection meeting, team members gathered and reviewed previous meetings prepared by the research team and discussed which development stage best characterized the team stage of each meeting.
The agreed-upon stage was then recorded as a meeting-level label.

\subsection{Annotated Dataset Summary}
\revise{After post-processing and annotation, the dataset contains 75,971 utterances from 88 dated meetings, represented in 104 transcript files, involving 50 recruited participants across 12 student teams.}
Team sizes ranged from 3 to 5 members; 7 teams primarily communicated in Korean and 5 in English.
Using post-meeting peer-perception surveys, we additionally assigned 352 emergent-role labels, with one label for each participant at each meeting for which survey responses were available.

The annotated interaction types were skewed toward a small number of frequent collaborative behaviors. \textit{Giving Information} was the most common category (35.29\%), followed by \textit{Active Listening} (13.09\%) and \textit{Linking Solutions} (12.19\%), suggesting that the meetings were dominated by task-relevant information sharing, solution-building, and responsive discussion.
Less frequent categories such as \textit{Task/Process Negative} (0.85\%), \textit{Social Negative} (0.61\%), and \textit{Linking \& Connecting} (0.11\%) appeared only rarely.
For emergent roles, \textit{Problem Solver} (17.61\%), \textit{Coordinator} (16.76\%), and \textit{Critic} (15.91\%) were the most common, indicating that task-oriented and coordination-oriented roles appeared more often than socially disruptive or off-task roles.
Detailed team-level statistics and full label distributions are provided in Appendix~\ref{appendix:dataset-stats}; annotated transcript excerpts are shown in Appendix~\ref{appendix:transcript-examples}.
\section{Experiments}
\label{sec:experiments}

We evaluate \dataset{} through two experiments, described below.

\subsection{Experimental Setup}
\label{sec:exp-setup}

\paragraph{Common configuration.}
Unless stated otherwise, we fine-tune with LoRA (rank~16, $\alpha$~=~32) on Gemma-3-12B-IT, trained on 32,243 samples \revise{(Teams~1,~2,~3,~5,~8,~9,~10,~11, and~12)}, validated on 1,857 samples (Team~4), and tested on 5,094 samples (Teams~6\&7) with 5-turn context windows.
\revise{Teams~6 and~7 together comprise approximately 13\% of all utterances, represent 3- and 4-member configurations, and span the full Tuckman lifecycle; holding them out also keeps the three largest teams in training. Team~4, the smallest team, is used for validation to minimize the amount of training data withheld.}
Fine-tuned models use completion-only loss; inference uses single-token logit scoring for open-source models and generative decoding (temperature~0) for proprietary APIs.
All results are single runs.
We additionally validate with Llama-3.1-8B-Instruct and benchmark four proprietary models (GPT-5.4, GPT-5.4-mini, Opus~4.6, Sonnet~4.6) in zero-shot.
All Korean transcripts were translated to English during post-processing (\S\ref{sec:postprocessing}); language labels (Korean/English) throughout refer to the language originally spoken during meetings.

We design two experiments:
\begin{enumerate}[leftmargin=*, itemsep=2pt]
\item \textbf{Prediction and analysis.} To probe whether models can learn group-level turn-taking patterns from \dataset{}, we evaluate next-speaker prediction (3--5~classes per team) and next-intention prediction (14 substantive classes, excluding \textit{Other/Neutral}) under three context conditions: \textbf{SU}~(Speaker + Utterance), \textbf{SRU}~(+ annotated role and Tuckman stage), and \textbf{S+R}~(structure only---speaker IDs, roles, and stages, with all utterance text removed).
To investigate how much team-specific data is required---a practical consideration for deploying such models to new teams---we further analyze \textit{data efficiency} by training single-team models with chronologically increasing meetings on Team~10 (Korean, 4~members) and Team~5 (English, 5~members).
To test whether \dataset{}-trained models generalize beyond our corpus, we conduct \textit{external validation} on the AMI Meeting Corpus~\citep{Carletta2005TheAM} (138 meetings, 4~English speakers each) using Llama-3.1-8B for direct comparison with \citet{hilgert-niehues-2025-next}, who report 47.85\% using AMI plus MultiLIGHT data.

\item \textbf{Generation.} To examine whether structural understanding translates to realistic utterance generation, we compare \revise{three conditions: \textbf{FT-Plain} (fine-tuned, plain generation: ``\textsc{Speaker}: utterance''), \textbf{FT-Reason} (fine-tuned with social-cue reasoning: the model generates an explicit role/intention prediction before producing the utterance), and \textbf{Vanilla} (Gemma-3-12B without fine-tuning)}, plus four proprietary models. \revise{Each generates 100 continuations for each generation length (1, 3, 5~turns).} We evaluate with automatic metrics (speaker accuracy, semantic similarity \revise{based on \texttt{all-MiniLM-L6-v2}}), a human evaluation on Prolific ($N$=66~evaluators, 1,142 pairwise judgments, \pounds4.50/evaluator), and LLM-as-judge evaluation.
\end{enumerate}

\subsection{Results}
\label{sec:results}

\paragraph{Experiment 1a: Speaker prediction.}
As a reference point, the bigram baseline---which predicts the most frequent next speaker given the current speaker---achieves 50.75\%.
Fine-tuning on \dataset{} yields large gains (Table~\ref{tab:prediction}): Gemma FT-SU reaches 64.53\%, +13.8\,pp over the bigram baseline.
Adding role labels helps minimally (FT-SRU 64.82\%, $\Delta$~=~+0.3\,pp), but removing text entirely still matches performance (FT-S+R 65.74\%), indicating that structural features alone are sufficient for predicting who speaks next.
This may reflect that utterance text introduces team-specific lexical patterns (e.g., project topics, jargon) that do not transfer to held-out teams, whereas structural features---speaker order, annotated roles, and development stages---capture more generalizable turn-taking regularities.
Fine-tuning outperforms all proprietary zero-shot models, with Gemma FT-SU (64.53\%) exceeding Opus~4.6 (63.25\%) and GPT-5.4 (61.97\%). Llama-3.1-8B replicates all patterns with comparable accuracy (Appendix~\ref{appendix:full-prediction}).
A longitudinal signal is also evident. Speaker prediction accuracy increases from 57.8\% in Forming-stage meetings to 70.3\% in Performing-stage meetings (+12.5\,pp; Appendix~\ref{appendix:detailed-breakdowns}), consistent with the theoretical expectation that established teams develop more predictable patterns.

\paragraph{Experiment 1b: Intention prediction.}
Intention prediction is a substantially harder task, likely due both to the larger number of classes and the subjective nature of categorization (Table~\ref{tab:prediction}).
The best model (FT-SU, 32.96\%) only marginally exceeds the majority baseline (30.05\%), and unlike speaker prediction, text is necessary (FT-S+R drops to 31.17\%).
No zero-shot model---proprietary or open-source---surpasses the majority baseline, consistent with \revise{the task's class imbalance} and annotation ambiguity.

\begin{table*}[t]
\centering
\small
\begin{tabular}{llcrrrr}
\toprule
& & & \multicolumn{2}{c}{\textbf{Speaker}} & \multicolumn{2}{c}{\textbf{Intention}} \\
\cmidrule(lr){4-5} \cmidrule(lr){6-7}
\textbf{Model} & \textbf{Context} & \textbf{Type} & \textbf{Acc.} & \textbf{$\Delta$ Bigram} & \textbf{Acc.} & \textbf{$\Delta$ Maj.} \\
\midrule
\multicolumn{7}{l}{\textit{Baselines}} \\
Random & --- & --- & 30.22 & --- & 7.14 & ---\\
Bigram & --- & --- & 50.75 & (ref) & --- & ---\\
Majority & --- & --- & --- & --- & 30.05 & (ref)\\
\midrule
\multicolumn{7}{l}{\textit{Zero-shot / Vanilla}} \\
Gemma-3-12B & SU & vanilla & 43.56 & $-$7.2 & 25.48 & $-$4.6 \\
\textit{GPT-5.4} & \textit{SU} & \textit{zero-shot} & \textit{61.97} & \textit{+11.2} & \textit{25.68} & \textit{$-$4.4} \\
\textit{Opus 4.6} & \textit{SU} & \textit{zero-shot} & \textit{63.25} & \textit{+12.5} & \textit{26.68} & \textit{$-$3.4} \\
\midrule
\multicolumn{7}{l}{\textit{Fine-tuned (Gemma-3-12B-IT + LoRA)}} \\
Gemma-3-12B & SU & FT & 64.53 & +13.8 & \textbf{32.96} & +2.9 \\
Gemma-3-12B & SRU & FT & 64.82 & +14.1 & 32.53 & +2.5 \\
Gemma-3-12B & S+R & FT & \textbf{65.74} & +15.0 & 31.17 & +1.1 \\
\bottomrule
\end{tabular}
\par\vspace{0.3em}
{\footnotesize Full results including Llama-3.1-8B, additional proprietary models, and few-shot baselines in Appendix~\ref{appendix:full-prediction}. Majority baseline for intention = ``Giving Information'' (30.05\%).}
\caption{Next-speaker and next-intention prediction on the \dataset{} test set (5,094 samples, Teams 6\&7). Context conditions: SU = Speaker + Utterance; SRU = + Role + Tuckman stage; S+R = structure only (no text). \textbf{Bold} = best. \textit{Italic} = proprietary zero-shot.}
\label{tab:prediction}
\end{table*}

\paragraph{Experiment 1c: Data efficiency.}
\label{result:1c}
\revise{Extending the prediction analysis to single-team settings, observed accuracy rises most sharply within the first three to four meetings (Table~\ref{tab:learning-curve}).
For Team~10, accuracy increases from 50.87\% with one meeting to 62.62\% with four; for Team~5, it increases from 50.00\% with one meeting to 63.84\% with three and then remains near 65\%.
Because the held-out set consists of the remaining meetings and therefore changes and shrinks as training meetings are added, these rows are not directly comparable as a fixed-test learning curve.
We interpret them as descriptive evidence that a small amount of team-specific data can recover much of the performance observed later in the sequence, a pattern also seen with Llama-3.1-8B (Appendix~\ref{appendix:lc-ami-details}).}

\begin{table*}[t]
\centering

\footnotesize
\renewcommand{\arraystretch}{0.92}
\setlength{\tabcolsep}{2.5pt}

\begin{tabular*}{\textwidth}{@{\extracolsep{\fill}}llllr@{}}
\toprule
\textbf{Team}
& \textbf{Meeting prefix}
& \textbf{Train $N$}
& \textbf{Acc. (\%)}
& \textbf{Ref. (\%)} \\
\midrule

Team 10 (Korean, $n$=4)
& 1/3/4/6
& 276/795/1,073/1,811
& 50.87/55.37/62.62/65.82
& 67.09 \\

Team 5 (English, $n$=5)
& 1/3/5/7
& 203/956/1,648/2,570
& 50.00/63.84/64.81/64.59
& 68.27 \\

\bottomrule
\end{tabular*}

\vspace{-0.45em}

\caption{
Chronological single-team adaptation with Gemma-3-12B.
Meeting prefixes, training sizes, and accuracies are listed in order.
Remaining meetings form the test set at each prefix, so results are
descriptive rather than fixed-test.}
\label{tab:learning-curve}

\vspace{3pt}

\footnotesize
\renewcommand{\arraystretch}{0.92}
\setlength{\tabcolsep}{5pt}

\begin{tabular*}{\textwidth}{@{\extracolsep{\fill}}lllr@{}}
\toprule
\textbf{Source}
& \textbf{Training condition}
& \textbf{Training data}
& \textbf{Acc. (\%)} \\
\midrule

H\&N
& Zero-shot Llama 8B
& ---
& 34.88 \\

H\&N
& Zero-shot Llama 70B
& ---
& 35.81 \\

Ours
& TIDES only (merged)
& 32K TIDES
& 35.33 \\

Ours
& TIDES only (unmerged)
& 62K TIDES
& 40.04 \\

Ours
& Balanced TIDES + AMI
& 124K total
& \textbf{45.79} \\

Ours
& Unbalanced TIDES + AMI
& 170K total
& 30.02 \\

H\&N
& Fine-tuned Llama 8B (in-domain)
& AMI + MultiLIGHT
& 47.85 \\

\bottomrule
\end{tabular*}

\vspace{-0.45em}

\caption{
AMI external validation (12,515 test samples; context window 8).
H\&N denotes \citet{hilgert-niehues-2025-next}.
``Unmerged'' preserves the original TIDES utterance boundaries;
the balanced mix uses approximately 42\% fewer examples than the
H\&N in-domain result
(Appendix~\ref{appendix:lc-ami-details}).
}
\label{tab:ami-validation}

\vspace{-0.5em}
\end{table*}

\paragraph{Experiment 1d: External validation.}
To test generalization beyond \dataset{}, we evaluate on AMI. A balanced TIDES+AMI mix reaches \textbf{45.79\%}, within 2.06\,pp of published SOTA with $\sim$42\% less training data (Table~\ref{tab:ami-validation}).
\revise{The TIDES-only transfer model (35.33\%) is close to Hilgert \& Niehues' zero-shot result (34.88\%), suggesting that some learned turn-taking regularities transfer across corpora.}
However, the full unbalanced mix (170K, trained for 1 epoch) degrades to 30.02\%, suggesting that data balance matters more than volume for cross-corpus transfer.

\paragraph{Experiment 2a: Generation: automatic metrics.}
FT-Reason achieves the highest 1-turn speaker accuracy (57.0\% vs.\ FT-Plain 49.0\%, Vanilla 14.0\%), and both FT conditions produce far fewer invalid speakers than Vanilla (152--178 vs.\ 786 at 10~turns), suggesting that fine-tuning on \dataset{} improves structural coherence in generation.
However, proprietary models outperform all fine-tuned models: Opus~4.6 reaches 62.0\% at 1-turn with zero invalid speakers, and degrades more slowly (39.0\% at 5-turn vs.\ FT-Reason's 29.2\%).
Proprietary models also show higher semantic similarity to ground truth (0.36--0.44 vs.\ 0.28--0.36; Table~\ref{tab:gen-auto}).
Unlike prediction, where fine-tuning closes the gap with proprietary models, generation quality appears to benefit more from model scale.

\paragraph{Experiment 2b: Generation: human evaluation.}
Despite the results above, human evaluators reveal a contrasting pattern (Table~\ref{tab:human-eval})\revise{: although most judgments involving Vanilla were ties (65.4--69.5\%), directional preferences \textbf{significantly favor Vanilla over fine-tuned outputs}}.
FT-Plain wins only 9.3\% of judgments vs.\ Vanilla's 21.1\% ($p<.001$); FT-Reason wins 12.7\% vs.\ 21.9\% ($p=.004$).
Vanilla scores 0.8--1.3 Likert points higher on naturalness, coherence, and speaker consistency ($d$~=~0.46--0.72; all $p$-values remain significant after Bonferroni correction).
FT outputs, while contextually grounded, exhibit surface artifacts (truncation, missing punctuation), whereas Vanilla generates polished but often off-topic continuations; however, the gap persisted even after we post-processed all FT outputs to correct capitalization, punctuation, and truncated endings (Appendix~\ref{appendix:generation-supplementary}).
\revise{This suggests that the difference extends beyond formatting: \textbf{improvements in structural modeling may not directly transfer to natural conversation generation}, as accurately predicting \textit{who} speaks next does not necessarily produce utterances that humans perceive as realistic.
We note, however, that our evaluators were not members of the recorded teams, and fine-tuning also adapts models to team-specific register and project-specific language; the current evaluation therefore cannot fully separate a structure--content mismatch from a domain-familiarity effect, and we treat this result as an open question.}

\begin{table*}[t]
\centering
\small
\begin{tabular}{lcccc}
\toprule
\textbf{Model} & \textbf{Preferred (\%)} & \textbf{Naturalness} & \textbf{Coherence} & \textbf{Speaker consistency} \\
\midrule
\multicolumn{5}{l}{\textit{FT-Plain vs. Vanilla} \quad (407 judgments; tie = 69.5\%)} \\
\quad FT-Plain & 9.3 & 3.07 & 2.78 & 3.05 \\
\quad Vanilla & \textbf{21.1}$^{***}$ & \textbf{3.99}$^{***}$ & \textbf{4.10}$^{***}$ & \textbf{3.93}$^{***}$ \\
\addlinespace[2pt]
\multicolumn{5}{l}{\textit{FT-Reason vs. Vanilla} \quad (347 judgments; tie = 65.4\%)} \\
\quad FT-Reason & 12.7 & 3.14 & 2.82 & 3.04 \\
\quad Vanilla & \textbf{21.9}$^{**}$ & \textbf{3.98}$^{***}$ & \textbf{4.06}$^{***}$ & \textbf{3.85}$^{***}$ \\
\addlinespace[2pt]
\multicolumn{5}{l}{\textit{FT-Plain vs. FT-Reason} \quad (388 judgments; tie = 53.4\%)} \\
\quad FT-Plain & 21.4 & 3.21 & 3.08 & 3.13 \\
\quad FT-Reason & \textbf{25.3} & \textbf{3.42} & \textbf{3.40}$^{***}$ & \textbf{3.41}$^{***}$ \\
\bottomrule
\end{tabular}
\par\vspace{0.3em}
{\scriptsize Preferred = percentage selecting that model; the remainder are ties. Ratings are mean 1--5 scores. Stars denote within-pair tests: $^{***}p<.001$, $^{**}p<.01$ after Bonferroni correction.}
\caption{Human evaluation (66 evaluators; 1,142 quality-controlled judgments). FT-Plain = direct fine-tuning; FT-Reason = social-cue fine-tuning; Vanilla = no fine-tuning.}
\label{tab:human-eval}
\end{table*}
\section{\revise{Limitations and Future Work}}

\dataset{} captures authentic social dynamics in Korean- and English-speaking team collaboration over the course of a semester, and therefore offers substantial potential for research directions that remain underexplored.

\paragraph{Alternative Utterance-level Annotation.}
Currently, \dataset{} adopts a modified version of \texttt{act4teams-SHORT} to capture the collaborative intent of each utterance.
Although the categories in \texttt{act4teams-SHORT} are well suited to representing utterances in collaborative settings, the relatively large number of categories and the inherently subjective nature of the annotation task contribute to only moderate inter-rater agreement.
In future work, we aim to extend the dataset with annotations based on more general dialogue-act taxonomies, such as AMI-DA~\citep{hain20072007} and MRDA~\citep{shriberg-etal-2004-icsi}.
Such annotations would improve the comparability of \dataset{} with prior work and facilitate its use in a broader range of downstream tasks.

\paragraph{Longitudinal Analysis.}
In experiment 1c (\S~\ref{result:1c}), we examined how much team-specific conversational data was needed for a model to adequately capture a team's interaction dynamics and predict the next speaker.
However, \dataset{} also includes longitudinal measures of collaboration satisfaction, emergent roles, and team development stages collected over the course of a semester, providing opportunities to track and analyze how a team's social dynamics evolve over time.
Accordingly, in future work, we aim to use \dataset{} to evaluate models' longitudinal capabilities, including predicting changes in members' emergent roles, transitions between team development stages, and changes in collaboration satisfaction from conversational histories.

\paragraph{Bilingual Data.}
\dataset{} includes both Korean and English data and provides English translations for teams that primarily communicated in Korean.
We also examined how the language composition of the training data affected next-speaker prediction, as described in Appendix~\ref{appendix:training-composition}.
In future work, we aim to extend this analysis by examining whether the evolution of within-team social dynamics differs across teams with different primary languages or cultural contexts.
\section{Conclusion}

We introduced \dataset{}, a longitudinal bilingual dataset tracking 12 university project teams over a full semester, with socio-structural annotations including emergent roles, interaction types, and team development stages.
Through a series of experiments, we demonstrated that fine-tuning on \dataset{} substantially improves next-speaker prediction---outperforming proprietary large-scale models---and that only a few hours of well-structured meeting data is sufficient for models to capture a team's interaction dynamics, with learned patterns transferring competitively to the AMI Meeting Corpus.

\revise{These findings indicate a potential mismatch between predicting conversational structure and producing content that users perceive as natural.}

Our experiments primarily evaluated local prediction within short context windows.
Modeling how team dynamics accumulate and shift across a team's full lifespan---for example, predicting role transitions, detecting phase shifts in real time, or adapting to evolving team norms---remains an important open direction.
The longitudinal structure and social annotations in \dataset{} open the door to such research, offering a resource where cross-meeting dynamics, role trajectories, and team development can be studied as they naturally unfold.

\section*{Acknowledgments}
\revise{This work was supported by the KAIST C2 (Creative \& Challenging) and UP Projects. This work was also supported by Institute of Information \& communications Technology Planning \& Evaluation (IITP) under the Leading Generative AI Human Resources Development (IITP-2026-RS-2026-25546560) grant funded by the Korea government (MSIT). We sincerely thank all participants for contributing their meeting data throughout the semester, our Prolific annotators and evaluators, and CSTL and KIXLAB members for insightful discussions and invaluable feedback.}

\section*{Ethics Statement}
This study was conducted under approval from our institute's Institutional Review Board (IRB). All participants provided informed consent prior to enrollment and were fully briefed on data collection procedures, including audio recording, during an orientation session. Participation was voluntary, and participants were free to withdraw at any time.

To protect participant privacy, only transcripts are made publicly available. Prior to any analysis or release, transcripts were processed through a multi-stage anonymization pipeline. Personally identifiable information---names, locations, and organizational references---was removed using Microsoft Presidio for English-speaking teams and a locally run Qwen3-30B-A3B model for Korean-speaking teams. All pseudonyms are gender-neutral. We compensated each team 500,000 KRW (approximately 325 USD) for their participation. Also, we compensated each participant 35,000 KRW per hour of reviewed audio if they validated the quality of transcripts.

For annotation on Prolific, annotators were compensated at 8 GBP (approximately 11 USD) per task (100 utterances), and for human evaluators, each evaluator was compensated at 4.5 GBP (approximately 6 USD) per task. Both rates exceeded the platform's recommended rate. There was no discrimination in the recruitment of annotators based on any demographic characteristics.

\bibliography{colm2026_conference, anthology-1, anthology-2}
\bibliographystyle{colm2026_conference}

\appendix
\section{Details for Utterance-level Interaction Type Annotation}
\label{appendix:utterance-level}

\subsection{Construction of Human-labeled Gold Data via Prolific}

\begin{figure}[ht]
    \centering
    \includegraphics[width=\linewidth]{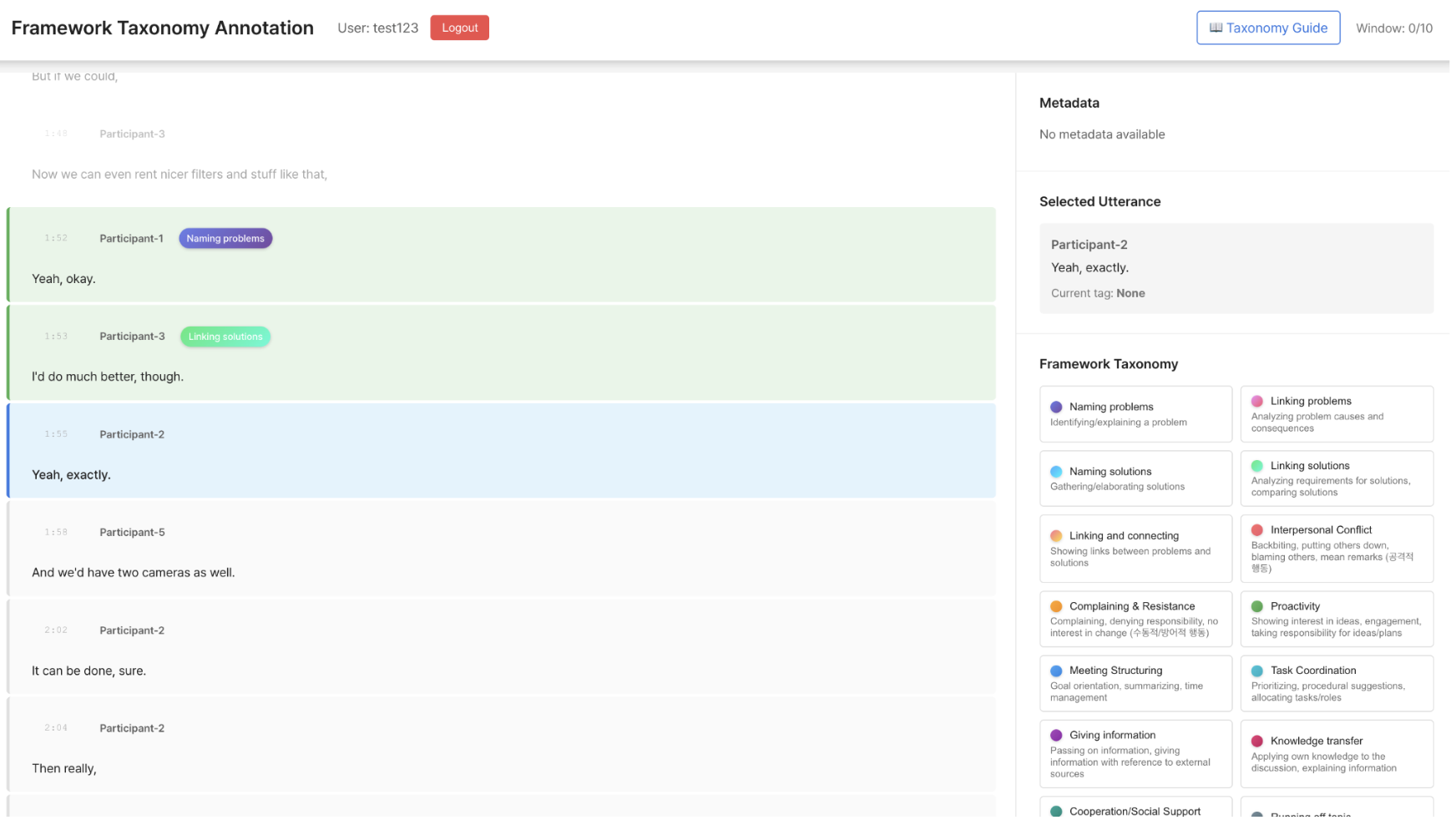}
    \caption{Custom annotation interface used for utterance-level interaction type annotation on Prolific. Annotators reviewed each target utterance with surrounding conversational context and selected one of the 15 modified \texttt{act4teams-SHORT} categories.}
    \label{fig:prolific-utterance-level}
\end{figure}

For utterance-level interaction type annotation, we recruited 260 annotators through the crowdsourcing platform Prolific.
Annotators accessed our custom-built annotation interface (Fig.~\ref{fig:prolific-utterance-level}), where they first provided their Prolific ID and then proceeded to the task.
Before starting the main annotation task, annotators were instructed to carefully review the definitions of the 15 categories in our modified \texttt{act4teams-SHORT} scheme.
To ensure sufficient understanding of the coding scheme, annotators were required to complete a tutorial quiz and answer all questions correctly before proceeding.
Each annotator was assigned a total of 100 utterances, organized into 10 windows of 10 target utterances each.
To help annotators interpret each utterance in context, we additionally provided the 10 preceding utterances and 10 following utterances for each window.
After completing all 10 windows, annotators received approximately 11 USD in compensation, subject to a quality check by the research team.

\subsection{Interaction Types in Human-labeled Gold Data}

\begin{table}[t]
\centering
\small
\begin{tabular}{lrr}
\toprule
Interaction Type & Count & Ratio (\%) \\
\midrule
Giving Information      & 1,881 & 32.97 \\
Active Listening        &   956 & 16.76 \\
Linking Solutions       &   395 &  6.92 \\
Other / Neutral         &   372 &  6.52 \\
Structuring             &   362 &  6.35 \\
Naming Solutions        &   339 &  5.94 \\
Social / Humor          &   296 &  5.19 \\
Proactivity             &   240 &  4.21 \\
Naming Problems         &   217 &  3.80 \\
Knowledge Transfer      &   171 &  3.00 \\
Cooperation             &   162 &  2.84 \\
Linking Problems        &   135 &  2.37 \\
Social Negative         &    66 &  1.16 \\
Task/Process Negative   &    62 &  1.09 \\
Linking \& Connecting   &    51 &  0.89 \\
\bottomrule
\end{tabular}
\caption{Distribution of utterance-level interaction types in the human-labeled gold data.}
\label{tab:gold-interaction-distribution}
\end{table}

Table~\ref{tab:gold-interaction-distribution} summarizes the distribution of utterance-level interaction types in the human-labeled gold data.
The distribution is imbalanced, with \textit{Giving Information} and \textit{Active Listening} accounting for a large proportion of the annotations, while several categories such as \textit{Social Negative}, \textit{Task/Process Negative}, and \textit{Linking \& Connecting} appear relatively infrequently.

\subsection{Utterance-Level Interaction Type Annotation}
We compared several base and fine-tuned models on the utterance-level interaction classification task using a leave-one-team-out validation setup (Table~\ref{tab:utterance-ft-mean}).
Although fine-tuned \texttt{Qwen3-14B} achieved the highest accuracy and Cohen’s $\kappa$, we selected the fine-tuned \texttt{Gemma-3-12B} for large-scale annotation because it achieved the best macro-F1 score (0.540), which we considered the most appropriate primary metric for our imbalanced multi-class setting.
The Fleiss’ $\kappa$ among human annotators was 0.400, indicating moderate agreement on this challenging 15-class annotation task.
\revise{The released corpus marks each utterance as either human-labeled gold or model-produced silver through the \texttt{annotation\_source} field. The final release contains 5,705 gold utterances and 70,266 silver utterances. Because the 15-way distribution is highly imbalanced, the silver layer should not be interpreted as uniformly reliable across categories; users can restrict analyses to the gold subset when higher-confidence supervision is required.}


\begin{table}[ht]
\centering
\small
\setlength{\tabcolsep}{5pt}
\renewcommand{\arraystretch}{1.05}
\begin{tabular}{lccc}
\toprule
Model & Acc. & $\kappa$ & F1 \\
\midrule
Llama3.1-8B (Base) & 0.382 & 0.286 & 0.231 \\
\rowcolor{gray!12}
Llama3.1-8B (FT)   & 0.687 & \underline{0.606} & \underline{0.528} \\
Qwen3-14B (Base)   & 0.488 & 0.397 & 0.313 \\
\rowcolor{gray!12}
Qwen3-14B (FT)     & \textbf{0.691} & \textbf{0.612} & 0.527 \\
Gemma-3-12B (Base) & 0.570 & 0.477 & 0.367 \\
\rowcolor{gray!12}
Gemma-3-12B (FT)   & \underline{0.687} & 0.605 & \textbf{0.540} \\
\bottomrule
\end{tabular}
\caption{Mean validation performance for utterance-level interaction type classification (leave-one-team-out). Best is bold; second-best is underlined.}
\label{tab:utterance-ft-mean}
\end{table}

\section{Annotated Dataset Statistics}
\label{appendix:dataset-stats}

\begin{table}[ht]
\centering
\small
\begin{minipage}[t]{0.5\linewidth}
\centering
\textbf{(a) Utterance-level interaction type distribution}

\vspace{0.4em}
\begin{tabular*}{\linewidth}{@{\extracolsep{\fill}}lrr@{}}
\toprule
Utterance Type & Count & Ratio (\%) \\
\midrule
Giving Information    & 26,810 & 35.29 \\
Active Listening      & 9,946  & 13.09 \\
Linking Solutions     & 9,260  & 12.19 \\
Naming Solutions      & 4,917  & 6.47  \\
Other / Neutral       & 4,703  & 6.19  \\
Structuring           & 4,643  & 6.11  \\
Proactivity           & 3,912  & 5.15  \\
Naming Problems       & 3,157  & 4.16  \\
Social / Humor        & 2,510  & 3.30  \\
Linking Problems      & 2,233  & 2.94  \\
Cooperation           & 1,812  & 2.39  \\
Knowledge Transfer    & 878  & 1.16  \\
Task/Process Negative & 642    & 0.85  \\
Social Negative       & 465    & 0.61  \\
Linking \& Connecting & 83     & 0.11  \\
\midrule
Total                 & 75,971 & 100.00 \\
\bottomrule
\end{tabular*}
\end{minipage}\hspace{0.05\linewidth}%
\begin{minipage}[t]{0.42\linewidth}
\centering
\textbf{(b) Emergent role distribution}

\vspace{0.4em}
\begin{tabular*}{\linewidth}{@{\extracolsep{\fill}}lrr@{}}
\toprule
Role & Count & Ratio (\%) \\
\midrule
Problem Solver   & 62 & 17.61 \\
Coordinator      & 59 & 16.76 \\
Critic           & 56 & 15.91 \\
Negative         & 38 & 10.80 \\
Task Motivator   & 32 & 9.09  \\
Team Leader      & 32 & 9.09  \\
Evaluator        & 16 & 4.55  \\
Attention Seeker & 13 & 3.69  \\
Task Completer   & 11 & 3.13  \\
Teamwork Support & 11 & 3.13  \\
Follower         & 10 & 2.84  \\
Power Seeker     & 7  & 1.99  \\
Social           & 5  & 1.42  \\
\midrule
Total            & 352 & 100.00 \\
\bottomrule
\end{tabular*}

\end{minipage}
\caption{Distribution of utterance-level interaction types and emergent roles in the dataset.}
\label{tab:data-distributions}
\end{table}

\begin{table}[ht]
\centering
\small
\begin{tabular}{lrr}
\toprule
Team & \# Utterances & \# Meetings \\
\midrule
Team\_3  & 14,058 & 8  \\
Team\_9  & 11,713 & 7  \\
Team\_2  & 11,630 & 13 \\
Team\_7  & 6,594  & 5  \\
Team\_5  & 6,168  & 10 \\
Team\_10 & 4,482  & 7  \\
Team\_12 & 4,000  & 12 \\
Team\_1  & 3,861  & 5  \\
Team\_11 & 3,852  & 6  \\
Team\_6  & 3,448  & 6  \\
Team\_8  & 3,377  & 5  \\
Team\_4  & 2,788  & 4  \\
\midrule
Total    & 75,971 & 88 \\
\bottomrule
\end{tabular}
\caption{Team-level statistics of the released transcript corpus. Meetings are counted by unique recording date; sessions split into multiple parts count as one meeting.}
\label{tab:team-stats}
\end{table}

The amount of conversation varied substantially across teams, ranging from 2,788 to 14,058 utterances per team, with an average of approximately 6,331 utterances per team.

\section{Transcript Examples}
\label{appendix:transcript-examples}

Tables~\ref{tab:example-korean} and~\ref{tab:example-english} show representative excerpts illustrating the full annotation layers of \dataset{}: pseudonymized speaker identity, emergent role, interaction type, and team development stage (in caption). Table~\ref{tab:example-korean} additionally shows Korean original utterances alongside their English translations.

\begin{table*}[ht]
\centering
\small
\setlength{\tabcolsep}{3pt}
\renewcommand{\arraystretch}{1.4}
\begin{tabular}{llp{3.5cm}p{4.1cm}l}
\toprule
\textbf{Speaker} & \textbf{Role} & \textbf{Korean (PII-removed)} & \textbf{English (translated)} & \textbf{Type} \\
\midrule
Alex (A) & Task Compl. & \begin{CJK}{UTF8}{mj}아 그러면 이제 두 분 중에 남는 분을 슬레이터로 맡기면 되는 건가요?\end{CJK} & Oh then, should we assign the remaining person to the slider role? & Naming Sol. \\
Blake (E) & Team Leader & \begin{CJK}{UTF8}{mj}네, 그래야 할 것 같아요.\end{CJK} & Yes, that seems like the way to go. & Active List. \\
Alex (A) & Task Compl. & \begin{CJK}{UTF8}{mj}저야 당연히 그러면 좋습니다.\end{CJK} & I'd be happy with that, of course. & Cooperation \\
Blake (E) & Team Leader & \begin{CJK}{UTF8}{mj}아니면 Drew님은 어떤 거 하는 게 편하세요?\end{CJK} & Or, Drew, which one would be more convenient for you? & Structuring \\
Drew (C) & Coordinator & \begin{CJK}{UTF8}{mj}어 저는 슬레이터가 더 좋아요.\end{CJK} & I prefer the slider. & Giving Info. \\
Blake (E) & Team Leader & \begin{CJK}{UTF8}{mj}그러면 맡길까요?\end{CJK} & Then shall we assign it to you? & Proactivity \\
\bottomrule
\end{tabular}
\caption{Transcript excerpt from Team~1 (Korean $\rightarrow$ English, meeting 2025-11-06). Team development stage: \textit{Performing}. The team is assigning filming roles for a video production project.}
\label{tab:example-korean}
\renewcommand{\arraystretch}{1.0}
\end{table*}

\begin{table*}[ht!]
\centering
\small
\setlength{\tabcolsep}{3pt}
\renewcommand{\arraystretch}{1.4}
\begin{tabular}{llp{7.5cm}l}
\toprule
\textbf{Speaker} & \textbf{Role} & \textbf{Utterance} & \textbf{Type} \\
\midrule
Riley (E) & Critic & I guess there is some random Academic Institute branch of the node pixel. & Giving Info. \\
Jordan (C) & Critic & Maybe you should make a copy of our stuff. & Naming Sol. \\
Morgan (A) & Negative & The LED is so bright that it can be fixed, but the hole is made here and there. & Linking Sol. \\
Morgan (A) & Negative & But to put several LEDs in a row, & Linking Sol. \\
Dakota (D) & Prob.\ Solver & I can't make the boards bend. & Naming Prob. \\
Dakota (D) & Prob.\ Solver & I think it's better to use glue gun or bond to connect them with LED lines. & Naming Sol. \\
\bottomrule
\end{tabular}
\caption{Transcript excerpt from Team~5 (English, meeting 2025-11-18). Team development stage: \textit{Norming}. The team is building a physical LED prototype for an HCI course project.}
\label{tab:example-english}
\renewcommand{\arraystretch}{1.0}
\end{table*}

\section{Meeting Metadata and Survey Examples}
\label{appendix:metadata-examples}

Table~\ref{tab:meeting-metadata-example} shows the meeting metadata for Team~5, illustrating how team development stages progress over the semester.
Table~\ref{tab:survey-example} shows post-meeting satisfaction scores and peer-evaluated emergent roles for the same team's first meeting.

\begin{table}[ht]
\centering
\small
\begin{tabular}{lcl}
\toprule
\textbf{Date} & \textbf{Stage} & \textbf{Topic (summary)} \\
\midrule
2025-10-28 & Forming & Brainstormed 5 ideas, narrowed to 3 \\
2025-11-03 & Forming & Finalized topic, planned implementation \\
2025-11-10 & Storming & Decided on circuits over Arduino \\
2025-11-18 & Norming & Realized neopixel implementation difficulty \\
2025-12-02 & Performing & Redistributed work after rack assembly \\
2025-12-06 & Performing & Bug fixing and integration \\
2025-12-10 & Performing & Button working, display planning \\
2025-12-12 & Performing & Product finalized and functional \\
2025-12-13 & Adjourning & Filmed demo video \\
2025-12-15 & Adjourning & Presentation role division \\
\bottomrule
\end{tabular}
\caption{Meeting metadata for Team~5 (English, 5 members). Stages were assigned retrospectively by team consensus via Tuckman self-assessment.}
\label{tab:meeting-metadata-example}
\end{table}

\begin{table}[ht!]
\centering
\small
\vspace{0.5em}
\setlength{\tabcolsep}{4pt}
\begin{tabular}{lcccccc}
\toprule
\textbf{Satisfaction Item} & \textbf{P1} & \textbf{P2} & \textbf{P3} & \textbf{P4} & \textbf{P5} & \textbf{Avg} \\
\midrule
Clear collaborative patterns & 3 & 3 & 3 & 4 & 5 & 3.6 \\
Members know their roles & 3 & 5 & 5 & 5 & 5 & 4.6 \\
Clear goals and working norms & 4 & 5 & 5 & 5 & 5 & 4.8 \\
Timely responses & 4 & 5 & 4 & 5 & 5 & 4.6 \\
Frequent communication & 5 & 5 & 4 & 5 & 5 & 4.8 \\
\bottomrule
\end{tabular}

\vspace{0.8em}

\begin{tabular}{lcccc}
\toprule
\textbf{Member} & \textbf{G1 (Task)} & \textbf{G2 (Social)} & \textbf{G3 (Domin.)} & \textbf{Assigned Role} \\
\midrule
Morgan (A) & 3.58 & 4.58 & 4.58 & Teamwork Support \\
Casey (B) & 4.33 & 4.89 & 4.89 & Coordinator \\
Jordan (C) & 4.44 & 4.11 & 4.22 & Power Seeker \\
Dakota (D) & 3.78 & 4.78 & 4.56 & Problem Solver \\
Riley (E) & 3.67 & 4.33 & 4.11 & Critic \\
\bottomrule
\end{tabular}
\caption{Post-meeting satisfaction survey and emergent roles for Team~5 (meeting 2025-10-28, stage: \textit{Forming}). Satisfaction items are rated 1--5; roles are derived from peer-evaluation scores across three TRIAD dimensions.}
\label{tab:survey-example}
\end{table}

\section{Full Post-Survey Question Set}
\label{appendix:full-survey}
To capture the evolution of group dynamics throughout the project, we administered a post-survey immediately after each meeting, as well as a final survey at the end of the project.
The post-survey consisted of three parts.
First, participants provided a brief reflection summarizing important decisions, turning points, or notable events from the meeting.
Second, they rated their satisfaction with team collaboration using questionnaire items adapted from prior work on collaboration and team processes~\cite{Ku2013CollaborationFT}.
Third, to capture emergent roles within the team, participants evaluated each of their teammates using nine peer-assessment items.
These items reflect the three dimensions of the TRIAD model—Dominance, Sociability, and Task Orientation—with three questions for each dimension~\citep{Driskell2017TeamRA}.
At the end of the semester, we additionally administered a final survey to assess participants’ overall satisfaction and perceived quality of the project outcome.
We also asked each team to collaboratively reflect on their meeting history and classify each meeting according to Tuckman’s Team Development Model—Forming, Storming, Norming, Performing, and Adjourning—along with a rationale for each classification~\citep{Tuckman1965DEVELOPMENTALSI}.

Below, we provide the full set of survey questions used in our study.

\subsection{Daily Meeting Reflection}
\begin{itemize}[leftmargin=*, itemsep=2pt]
    \item Please share the important decision or turning point during the conversation of today's meeting.
\end{itemize}

\subsection{Team Collaboration Satisfaction}
\textit{Note: All items are measured on a 5-point Likert scale (1 = Strongly Disagree to 5 = Strongly Agree).}
\begin{enumerate}[leftmargin=*, itemsep=2pt]
    \item My team develops clear collaborative patterns to increase team learning efficiency.
    \item My team members clearly know their roles during the collaboration.
    \item My team has an efficient way to track the edition of documents.
    \item My team sets clear goals and establishes working norms.
    \item My team members reply to all responses in a timely manner.
    \item My team members communicate with each other frequently.
    \item I trust each team member can complete his/her work on time.
    \item My team is receiving feedback from each other.
    \item Communicating with team members regularly helps me to understand the team project better.
    \item My team members encourage open communication with each other.
    \item My team members communicate in a courteous tone.
\end{enumerate}

\subsection{Emergent Role Survey (Peer Evaluation)}
\textit{Note: These questions evaluate the behavior of each team member on a 5-point frequency scale (1 = Never to 5 = Always).}
\begin{enumerate}[leftmargin=*, itemsep=2pt]
    \item Did this team member actively lead group discussions?
    \item Did this team member clearly present the team’s activities or direction?
    \item Did this team member assert their opinions in decision-making and influence the team?
    \item Did this team member create a positive and comfortable atmosphere in the team?
    \item Did this team member respect and support other team members’ feelings and opinions?
    \item Did this team member help mediate or facilitate smooth communication during conflicts?
    \item Did this team member help the team stay focused on its shared goals and avoid distractions?
    \item Did this team member fulfill their assigned role responsibly and on time?
    \item Did this team member show a diligent and meticulous attitude to improve task quality?
\end{enumerate}

\subsection{Final Survey (Individual Assessment)}
\begin{enumerate}[leftmargin=*, itemsep=2pt]
    \item How satisfied are you overall with the project results?
    \item Do you think the project has achieved its intended outcome?
    \item How do you rate the quality of the project results?
\end{enumerate}

\subsection{Final Survey (Team Reflection)}
\begin{itemize}[leftmargin=*, itemsep=2pt]
    \item \textbf{Group Discussion Task:} Looking back at all previous meetings, discuss with your team members and assign each meeting to the corresponding stage of Tuckman's Team Development Model (Forming, Storming, Norming, Performing, Adjourning). Please provide the rationale for your classifications.
\end{itemize}


\section{Post-processing Pipeline Details}
\label{appendix:postprocessing}

This appendix provides full technical details for the six-stage post-processing pipeline described in Section~\ref{sec:postprocessing}.
The pipeline converts raw audio into privacy-preserved, speaker-identified transcripts through the following stages:
(1)~ASR, (2)~speaker diarization, (3)~PII removal, (4)~translation, (5)~cross-session speaker unification, and (6)~utterance boundary refinement.

\subsection{Stage 1: ASR}

We transcribed all audio files using Whisper Large-V3~\citep{radford2023robust} with float16 precision on a single RTX A6000 GPU.
Table~\ref{tab:asr-config} summarizes the configuration.
Audio files were first preprocessed with FFmpeg to remove silence regions using the filter \texttt{silenceremove=1:0:-50dB} and resampled to 16\,kHz mono.
We loaded Whisper Large-V3 via the HuggingFace \texttt{transformers} pipeline with hallucination reduction enabled through a repetition penalty of 1.1 and \texttt{no\_repeat\_ngram\_size} of 3.

\begin{table}[ht]
\centering
\small
\begin{tabular}{ll}
\toprule
Parameter & Value \\
\midrule
Model & \texttt{openai/whisper-large-v3} \\
Device / Precision & CUDA / float16 \\
Batch size & 24 \\
Temperature & 0.2 \\
Entropy threshold & 2.8 \\
Sample rate & 16,000\,Hz \\
Repetition penalty & 1.1 \\
No-repeat n-gram size & 3 \\
Return timestamps & Segment-level \\
\bottomrule
\end{tabular}
\caption{ASR configuration for Whisper Large-V3.}
\label{tab:asr-config}
\end{table}

\subsection{Stage 2: Speaker Diarization}

Raw Whisper output does not distinguish speakers.
We developed a \textit{Smart Pipeline} that combines pyannote 3.1~\citep{bredin2023pyannote,plaquet2023powerset} for voice activity detection with ECAPA-TDNN~\citep{desplanques2020ecapa} speaker embeddings (192-dim) and Median Absolute Deviation outlier filtering, then re-clusters to the known team size $N$, ensuring correct speaker counts by construction.

\paragraph{\textit{Smart Pipeline}.}
The pipeline proceeds in five steps:
\begin{enumerate}[leftmargin=*, itemsep=2pt]
    \item \textbf{VAD:} pyannote 3.1 (\texttt{pyannote/speaker-diarization-3.1}) segments the audio into speech and non-speech regions.
    \item \textbf{Segment extraction:} Speech regions are extracted as individual audio segments.
    \item \textbf{Embedding extraction:} Each segment is encoded with ECAPA-TDNN (\texttt{speechbrain/spkrec-ecapa-voxceleb}), producing a 192-dimensional speaker embedding. Segments shorter than 0.5\,s are zero-padded.
    \item \textbf{MAD filtering:} Speaker-level embeddings are computed by averaging segment embeddings per speaker, with Median Absolute Deviation filtering to remove outlier segments.
    \item \textbf{Re-clustering:} Embeddings are re-clustered to the known team size $N$ using multiple methods (KMeans, Agglomerative with ward/complete/average linkage, Spectral clustering), and the best result is selected.
\end{enumerate}

\subsection{Stage 3: PII Removal}

English and Korean require fundamentally different PII removal approaches.
For English, we used Microsoft Presidio~\citep{presidio}, a rule-based detection framework with entity linking and consistent pseudonymization.
For Korean, we employed Qwen3-30B-A3B-Instruct-2507~\citep{qwen3} running locally as a context-aware detector; an earlier regex-based approach had yielded a 91\% false-positive rate, which the LLM-based pipeline substantially reduced, though 375 manual corrections were still needed.
All replacements use gender-neutral pseudonyms (e.g., Alex, Jordan, Taylor).

\paragraph{English Pipeline.}
For the five English/mixed teams (329,537 segments), we used Microsoft Presidio with three custom components:
\begin{itemize}[leftmargin=*, itemsep=2pt]
    \item \textbf{EntityLinker:} Maps name variants (e.g., ``Mary,'' ``MJ,'' ``Jane'') to canonical forms.
    \item \textbf{ConsistentAnonymizer:} Ensures the same entity always maps to the same pseudonym across the entire dataset.
    \item \textbf{PresidioPIIMaskerV2:} Wraps the Presidio \texttt{AnalyzerEngine} with custom Korean phone number and ID recognizers.
\end{itemize}
Detected entity types include: \textsc{person}, \textsc{location}, \textsc{organization}, \textsc{email}, \textsc{phone\_number}, \textsc{url}, and \textsc{date\_time}.

\paragraph{Korean Pipeline.}
For the seven Korean teams (48,697 segments), we employed Qwen3-30B-A3B-Instruct-2507 running locally (\texttt{bfloat16}, \texttt{device\_map=auto}) in a three-stage process:
\begin{enumerate}[leftmargin=*, itemsep=2pt]
    \item \textbf{Detection:} The LLM identifies \textsc{person}, \textsc{location}, and \textsc{organization} entities directly from Korean text, generating an \texttt{entity\_mapping.json}.
    \item \textbf{Replacement:} Detected entities are replaced using regex patterns with Korean-aware word boundary matching (lookbehind for Hangul/ASCII, lookahead for ASCII only, since Korean particles attach directly to names).
    \item \textbf{Translation:} PII-replaced Korean text is translated to English by the same Qwen3 model, running as a subprocess to ensure GPU memory cleanup between files.
\end{enumerate}

\paragraph{Pseudonym Pools.}
\begin{itemize}[leftmargin=*, itemsep=2pt]
    \item \textbf{Person:} 32 gender-neutral English names (Alex, Jordan, Taylor, Morgan, Casey, Riley, Quinn, Avery, Parker, Drew, Reese, Jamie, Sage, River, Phoenix, Blake, Charlie, Emerson, Hayden, Skyler, Dakota, Finley, Rowan, Ellis, Cameron, Peyton, Logan, Spencer, Bailey, Kendall, Harper, Addison).
    \item \textbf{Location:} Coded patterns (Building-A, Room-101, Campus-North, City-A, Lab-Alpha, etc.).
    \item \textbf{Organization:} Coded patterns (Tech-Lab, Company-A, Institute-Alpha, etc.).
\end{itemize}

\paragraph{Quality Assurance.}
After automated PII removal, we ran a PII leak checker across all anonymized files (TXT, CSV, XLSX formats) to detect residual personal information.
The Korean pipeline required many manual corrections, primarily due to common Korean names that are also everyday words (e.g., some names are identical to common pronouns).

\subsection{Stage 4: Translation}

To unify the dataset into a single language for downstream modeling, we translated all Korean transcripts to English using Qwen3-30B-A3B-Instruct-2507, running entirely on local machines.

All Korean transcripts were translated to English using Qwen3-30B-A3B-Instruct-2507 (\texttt{bfloat16}, \texttt{device\_map=auto}).
Translation was performed per-file in isolated subprocesses to ensure complete GPU memory release between files.
Segments without Korean characters were automatically skipped via a \texttt{has\_korean()} check.
For segments exceeding 200 characters, individual (rather than batch) translation was used to maintain quality.
Pseudonym preservation was validated post-translation using the \texttt{validate\_pseudonyms()} function, which verifies that all pseudonyms present in the source appear in the translated output.

\subsection{Stage 5: Cross-session Speaker Unification}

Diarization assigns arbitrary speaker labels that reset across sessions, so the same individual may receive different labels in different recordings.
We unified identities using WeSpeaker~\citep{wang2022wespeaker} 256-dimensional embeddings with the Hungarian algorithm~\citep{kuhn1955hungarian} and a complementary \textit{mention matrix}---leveraging the observation that speakers rarely say their own name---producing 422 speaker-file mappings with a within-speaker cosine similarity of 0.89 versus 0.34 between speakers.

\paragraph{Step 1: Embedding Extraction \& Hungarian Matching.}
We used the WeSpeaker model (\texttt{pyannote/wespeaker-voxceleb-resnet34-LM}) to extract 256-dimensional speaker embeddings.
For each speaker in each session, up to 5 segments were randomly sampled (minimum duration 1.5\,s, capped at 30\,s), and their embeddings were averaged to obtain a representative centroid.
Files were processed in order of decreasing speaker count: the first file initializes the centroid pool, and subsequent files are matched against existing centroids using the Hungarian algorithm with cost matrix $C = 1 - \text{cosine\_similarity}(E, M)$, where $E$ is the embedding matrix and $M$ is the centroid matrix.
Unmatched speakers create new centroids (e.g., \texttt{PERSON\_F}).

\paragraph{Step 2: Label Application.}
The resulting CSV mapping (422 entries) was applied to update the \texttt{speaker} field in all released JSON files, converting arbitrary per-session labels (\texttt{PERSON\_0}, \texttt{PERSON\_1}, \ldots) to consistent cross-session labels (\texttt{PERSON\_A}, \texttt{PERSON\_B}, \ldots).
Team~1 was mapped manually and verified independently.

\paragraph{Step 3: Mention Matrix Inference.}
As a complementary identity signal, we counted how often each speaker mentions each pseudonym using word-boundary regex matching, leveraging the observation that speakers rarely say their own name.
Each assignment received a confidence tag:
\begin{itemize}[leftmargin=*, itemsep=2pt]
    \item \textbf{VERIFIED:} Manual verification (Team~1 only).
    \item \textbf{HIGH:} 0 self-mentions and $\geq$3 total mentions by others.
    \item \textbf{MED:} $\leq$1 self-mention and $\geq$2 total mentions.
    \item \textbf{LOW:} All other cases.
\end{itemize}
This produced 49 speaker-to-real-identity mappings across 12 teams.

\subsection{Stage 6: Utterance Boundary Refinement}

We refined utterance boundaries using GPT-5-mini via the OpenAI API, selected for its reliable structured JSON output and applied only to already-anonymized transcripts.
The model decides one of four actions per utterance---\textsc{keep}, \textsc{merge\_next}, \textsc{merge\_prev}, or \textsc{split}---guided by 10 rules from the Language Development Project (LDP) transcription guidelines~\citep{macwhinney2000childes}, using a sliding window of 8 utterances.
Hard constraints prevent merging across different speakers or pauses $\geq$\,2\,s.

\paragraph{Model \& Configuration.}
We used \texttt{gpt-5-mini} via the OpenAI API (\texttt{AsyncOpenAI}) with the following settings:

\begin{table}[ht]
\centering
\small
\begin{tabular}{ll}
\toprule
Parameter & Value \\
\midrule
Model & \texttt{gpt-5-mini} \\
Window size & 8 utterances \\
Stride & 6 (overlap = 2) \\
Pause threshold & 2.0\,s \\
Max concurrent calls & 10 \\
Max retries & 3 (exponential backoff) \\
\bottomrule
\end{tabular}
\caption{Utterance boundary refinement configuration.}
\label{tab:refinement-config}
\end{table}

\paragraph{LDP Rules.}
The system encodes 10 numbered rules from the Language Development Project transcription guidelines, plus 3 unnumbered heuristic rules (13 prompt items total).
Two rules serve as hard constraints that cannot be overridden:

\begin{itemize}[leftmargin=*, itemsep=2pt]
    \item \textbf{Rule 4.6:} Never merge utterances from different speakers.
    \item \textbf{Rule 4.8.1:} Never merge across pauses $\geq$\,2\,seconds.
\end{itemize}

The remaining rules govern splitting and merging decisions:

\begin{table}[ht]
\centering
\small
\begin{tabular}{llll}
\toprule
Rule & Category & Action & Description \\
\midrule
4.3 & Self-correction & Keep & Within-thought corrections \\
4.4 & Abandoned thought & Split & Different-thought restart \\
4.5 & False start & Keep & Same topic, same utterance \\
4.7 & Multi-sentence & Split & No conjunction between sentences \\
4.9.1 & Non-sentence tag & Merge & ``ok,'' ``right,'' ``honey'' \\
4.9.3 & Complete tag & Split & ``I know,'' ``I don't know'' \\
4.9.4 & Perception verb & Keep & ``I know it's crazy'' = 1 utt. \\
4.10 & Repetition & Keep & ``no no no'' preserved \\
\bottomrule
\end{tabular}
\caption{LDP rules encoded in the utterance boundary refinement system.}
\label{tab:ldp-rules}
\end{table}

\paragraph{Processing Results.}
The released corpus contains 104 transcript files from 12 teams spanning 88 unique meeting dates; longer sessions were split into multiple parts. Across the processed corpus:
\begin{itemize}[leftmargin=*, itemsep=2pt]
    \item Splits applied: 1,238 (including multi-sentence splits producing 2--5 segments each); Merges applied: 309
    \item API calls: 1,015 windows processed
    \item Error rate: 0\% (all windows successful)
\end{itemize}

\section{Full Prediction Results}
\label{appendix:full-prediction}

Table~\ref{tab:prediction-full} shows the complete results for Experiment~1, including all proprietary models, few-shot baselines, and Llama-3.1-8B fine-tuned conditions.

\begin{table*}[ht]
\centering
\small
\begin{tabular}{llcrrrr}
\toprule
& & & \multicolumn{2}{c}{\textbf{Speaker}} & \multicolumn{2}{c}{\textbf{Intention}} \\
\cmidrule(lr){4-5} \cmidrule(lr){6-7}
\textbf{Model} & \textbf{Context} & \textbf{Type} & \textbf{Acc.} & \textbf{$\Delta$ Bigram} & \textbf{Acc.} & \textbf{$\Delta$ Maj.} \\
\midrule
\multicolumn{7}{l}{\textit{Baselines}} \\
Random & --- & --- & 30.22 & --- & 7.14 & ---\\
Bigram & --- & --- & 50.75 & (ref) & --- & ---\\
Majority & --- & --- & --- & --- & 30.05 & (ref)\\
\midrule
\multicolumn{7}{l}{\textit{Zero-shot / Vanilla}} \\
Llama-3.1-8B & SU & vanilla & 29.07 & $-$21.7 & 15.57 & $-$14.5 \\
Gemma-3-12B & SU & vanilla & 43.56 & $-$7.2 & 25.48 & $-$4.6 \\
\textit{GPT-5.4-mini} & \textit{SU} & \textit{zero-shot} & \textit{59.23} & \textit{+8.5} & \textit{24.22} & \textit{$-$5.8} \\
\textit{Sonnet 4.6} & \textit{SU} & \textit{zero-shot} & \textit{59.70} & \textit{+9.0} & \textit{25.38} & \textit{$-$4.7} \\
\textit{GPT-5.4} & \textit{SU} & \textit{zero-shot} & \textit{61.97} & \textit{+11.2} & \textit{25.68} & \textit{$-$4.4} \\
\textit{Opus 4.6}$^\dagger$ & \textit{SU} & \textit{zero-shot} & \textit{63.25} & \textit{+12.5} & \textit{26.68} & \textit{$-$3.4} \\
Gemma-3-12B & SU & 7-shot & 53.85 & +3.1 & --- & --- \\
\midrule
\multicolumn{7}{l}{\textit{Fine-tuned (Gemma-3-12B-IT + LoRA)}} \\
Gemma-3-12B & SU & FT & 64.53 & +13.8 & \textbf{32.96} & +2.9 \\
Gemma-3-12B & SRU & FT & 64.82 & +14.1 & 32.53 & +2.5 \\
Gemma-3-12B & S+R & FT & \textbf{65.74} & +15.0 & 31.17 & +1.1 \\
\midrule
\multicolumn{7}{l}{\textit{Fine-tuned (Llama-3.1-8B + LoRA)}} \\
Llama-3.1-8B & SU & FT & 65.17 & +14.4 & \textbf{33.18} & +3.1 \\
Llama-3.1-8B & SRU & FT & \textbf{66.20} & +15.5 & 32.82 & +2.8 \\
Llama-3.1-8B & S+R & FT & 65.65 & +14.9 & 31.72 & +1.7 \\
\bottomrule
\end{tabular}
\par\vspace{0.3em}
{\footnotesize $^\dagger$Speaker: 154 API errors (3.0\%); Intention: 57 errors (1.1\%), treated as incorrect. Retrying yields 65.04\% speaker accuracy, above FT-SU (64.53\%) but below the best fine-tuned condition (S+R, 65.74\%). Majority baseline for intention = ``Giving Information'' (30.05\%). 7-shot intention not evaluated.}
\caption{Full prediction results for Experiment~1, including all models and conditions. See Table~\ref{tab:prediction} for the main results.}
\label{tab:prediction-full}
\end{table*}

\section{Model-Agnostic Validation}
\label{appendix:model-agnostic}

To confirm that our findings are not artifacts of a specific model, we replicate the full Experiment~1 conditions with Llama-3.1-8B-Instruct using identical LoRA configuration (rank~16, $\alpha$~=~32, 2~epochs, lr~=~2e-4).
Table~\ref{tab:llama-full} shows results across all six conditions.

\begin{table}[ht]
\centering
\small
\makebox[\linewidth][c]{%
\begin{tabular}{llcccc}
\toprule
& & \multicolumn{2}{c}{\textbf{Speaker (\%)}} & \multicolumn{2}{c}{\textbf{Intention (\%)}} \\
\cmidrule(lr){3-4} \cmidrule(lr){5-6}
\textbf{Format} & \textbf{Mode} & Gemma & Llama & Gemma & Llama \\
\midrule
SU & Vanilla & 43.56 & 29.07 & 25.48 & 15.57 \\
SU & FT & 64.53 & 65.17 & 32.96 & 33.18 \\
SRU & Vanilla & 38.69 & 28.01 & 24.48 & 14.96 \\
SRU & FT & 64.82 & 66.20 & 32.53 & 32.82 \\
S+R & Vanilla & 47.78 & 42.64 & 21.50 & 11.17 \\
S+R & FT & 65.74 & 65.65 & 31.17 & 31.72 \\
\bottomrule
\end{tabular}%
}
\vspace{0.3em}
\newline
{\footnotesize All FT results are within 0.1--1.4\,pp across models for speaker prediction, and within 0.2--0.6\,pp for intention prediction, suggesting model-agnostic patterns.}
\caption{Model-agnostic validation: Gemma-3-12B vs.\ Llama-3.1-8B across all conditions. Same LoRA config, same data splits.}
\label{tab:llama-full}
\end{table}

\section{Learning Curve and AMI Experiment Details}
\label{appendix:lc-ami-details}

\paragraph{Learning curve setup.}
For each team, we chronologically order all meetings and incrementally add them to the training set. 
At step $k$, the model trains on meetings 1 through $k$ and is tested on all remaining meetings $k{+}1$ through $N$.
This means that the test set changes and shrinks as $k$ increases, so rows should not be interpreted as evaluations on an identical fixed test set; late-stage estimates also have high variance and may exceed the cross-team reference.
Training uses the same LoRA configuration as Experiment~1, with the SU (Speaker + Utterance) format.
We additionally evaluate the 9-team general model (trained on all 32,243 samples) on each team's full test data to establish a cross-team reference.
Table~\ref{tab:learning-curve-full} reports the full learning-curve results for both Gemma-3-12B and Llama-3.1-8B; the main paper (Table~\ref{tab:learning-curve}) reports Gemma only.

\begin{table}[ht]
\centering
\small
\setlength{\tabcolsep}{4pt}
\begin{tabular}{cccccc}
\toprule
& & \multicolumn{2}{c}{\textbf{Gemma-3-12B}} & \multicolumn{2}{c}{\textbf{Llama-3.1-8B}} \\
\cmidrule(lr){3-4} \cmidrule(lr){5-6}
\textbf{Mtgs} & \textbf{Train $N$} & \textbf{Acc.} & \textbf{\% ref.} & \textbf{Acc.} & \textbf{\% ref.} \\
\midrule
\multicolumn{6}{l}{\textit{Team 10 (Korean, 4 members, 7 meetings)}} \\
1 & 276 & 50.87\% & 75.8\% & 53.61\% & 75.6\% \\
3 & 795 & 55.37\% & 82.5\% & 58.46\% & 82.4\% \\
4 & 1,073 & 62.62\% & 93.3\% & 63.20\% & 89.2\% \\
6 & 1,811 & 65.82\% & 98.1\% & 64.35\% & 90.8\% \\
\midrule
\multicolumn{6}{l}{\textit{Team 5 (English, 5 members, 10 meetings)}} \\
1 & 203 & 50.00\% & 73.2\% & 60.73\% & 99.6\% \\
3 & 956 & 63.84\% & 93.5\% & 64.83\% & 106.3\% \\
5 & 1,648 & 64.81\% & 94.9\% & 65.10\% & 106.8\% \\
7 & 2,570 & 64.59\% & 94.6\% & 64.68\% & 106.1\% \\
\bottomrule
\end{tabular}
\par\vspace{0.3em}
{\footnotesize References---Gemma: 67.09\% (T10), 68.27\% (T5); Llama: 70.89\% (T10), 60.96\% (T5). Llama T5 exceeds its reference due to the small test set (187~samples).}
\caption{Full chronological single-team adaptation results for both models. \% ref.\ is relative to each model's 9-team general model evaluated on that team; because the remaining-meeting test set changes across rows, the percentages are descriptive and are not fixed-test learning-curve estimates.}
\label{tab:learning-curve-full}
\end{table}

\paragraph{AMI data preparation.}
We use the AMI Meeting Corpus from HuggingFace (\texttt{edinburghcstr/ami}, IHM configuration).
The official test split contains 16~meetings from 4~groups (EN2002, ES2004, IS1009, TS3003).
We construct sliding-window samples with context-size~8 to match \citet{hilgert-niehues-2025-next}, yielding 12,515~test samples.
Speakers are anonymized to PERSON\_A/B/C/D by order of first appearance in each meeting.

For the balanced TIDES+AMI training mix (124K), we use TIDES in unmerged format (62,221~samples, context~8) and downsample AMI training data from $\sim$108K to 62,221~samples to match.
The unmerged format preserves finer utterance boundaries, that align with AMI's segmentation style.
The full unbalanced mix (170,038 samples, trained for 1 epoch) uses all available data from both corpora without downsampling, but degrades accuracy to 30.02\%---below both the TIDES-only transfer condition (35.33\%) and Hilgert \& Niehues' zero-shot result (34.88\%)---suggesting that corpus balance is important for cross-corpus transfer.

\section{Training Composition and Cross-Culture Analysis}
\label{appendix:training-composition}

To understand how language composition in training data affects speaker prediction, we fix the test set (Teams~6\&7, Korean) and vary training composition across four conditions: Korean-only, English-only, balanced (Korean downsampled to match English), and the original mixed set. Note that ``Korean teams'' and ``English teams'' refer to the language originally spoken during recorded meetings, not the actual language of the post-processed data.

\begin{table}[ht]
\centering
\small
\begin{tabular}{lcrr}
\toprule
\textbf{Condition} & \textbf{Lang.} & \textbf{Train $N$} & \textbf{Acc. (\%)} \\
\midrule
Vanilla & --- & 0 & 43.56 \\
Bigram & --- & --- & 50.75 \\
\midrule
EN-only & 100\% EN & 8,703 & 57.20 \\
Balanced & 50/50 KR/EN & 17,406 & 58.74 \\
KR-only & 100\% KR & 23,540 & 60.21 \\
Mixed (orig.) & 73/27 KR/EN & 32,243 & \textbf{64.53} \\
\bottomrule
\end{tabular}
\caption{Effect of training language composition on speaker prediction. Test set fixed: Teams 6\&7 (Korean). Gemma-3-12B-IT + LoRA.}
\label{tab:training-composition}
\end{table}

Table~\ref{tab:training-composition} shows that \textbf{data volume is the primary driver}: Mixed (64.53\%, 32K samples) outperforms all alternatives, and KR-only (60.21\%) surpasses EN-only (57.20\%) by 3.0\,pp, reflecting a same-language advantage for the Korean test set.
EN-only still exceeds the bigram baseline by +6.4\,pp, suggesting that cross-lingual transfer of turn-taking dynamics is substantial---conversation structure transfers across languages even without target-language data.
Balanced (58.74\%) slightly exceeds EN-only but falls below KR-only, indicating that subsampling from 23K to 8.7K Korean samples carries a cost not fully offset by language diversity.

\section{Detailed Breakdowns}
\label{appendix:detailed-breakdowns}

\paragraph{Per-role analysis.}
Speaker prediction accuracy varies by the target speaker's emergent role.
Attention Seekers are most predictable (75.4\% on the test set), likely because their frequent backchanneling creates strong sequential patterns.
Critics are hardest to predict (53.1\%), consistent with their tendency to interject at unpredictable moments.

\paragraph{Tuckman stage progression.}
Teams in the Forming stage (Tuckman) show lower speaker prediction accuracy (57.8\% on the test set) than teams in the Performing stage (70.3\%), a +12.5\,pp gap.
This aligns with theory: established teams develop more predictable interaction patterns as they mature.

\paragraph{Intention class analysis.}
The fine-tuned model's intention predictions concentrate on Giving Information (74\%) and Active Listening (21\%), achieving top-3 accuracy of 62.6\% but macro F1 of only 0.072.
\revise{This class collapse reflects the skewed label distribution and ambiguity of the fine-grained categories, suggesting that intention prediction may require coarser targets, richer contextual signals, or task-specific architectures.}

\section{Generation Supplementary}
\label{appendix:generation-supplementary}

\paragraph{Automatic evaluation.}
Table~\ref{tab:gen-auto} shows that FT-Reason consistently outperforms FT-Plain on structural metrics (speaker accuracy, ROUGE-L) while Vanilla produces the most invalid speakers (786 at 10-turn vs.\ 152--178 for FT).
Degeneration rates increase with generation length: 1\% at 1-turn, 9\% at 3-turn, and 23\% at 5-turn, motivating the 4-stage quality filtering pipeline described below.

\begin{table*}[t]
\centering
\small
\setlength{\tabcolsep}{4pt}
\begin{tabular}{lrrrrcrrr}
\toprule
& \multicolumn{3}{c}{\textbf{Speaker Acc. (\%)}} & \textbf{ROUGE-L} & \textbf{Inv.} & \multicolumn{3}{c}{\textbf{Semantic Sim.}} \\
\cmidrule(lr){2-4} \cmidrule(lr){7-9}
\textbf{Condition} & 1t & 3t & 5t & (1t) & Spkrs$^\dagger$ & 1t & 3t & 5t \\
\midrule
\multicolumn{9}{l}{\textit{Fine-tuned (Gemma-3-12B)}} \\
FT-Plain (plain generation)      & 49.0 & 31.0 & 28.6 & 4.79 & 178 & .307 & .321 & .345 \\
FT-Reason (social-cue reasoning) & 57.0 & 37.7 & 29.2 & 5.46 & 152 & .283 & .317 & .360 \\
Vanilla (no fine-tuning)         & 14.0 & 21.7 & 20.2 & 4.41 & 786 & .319 & .346 & .348 \\
\midrule
\multicolumn{9}{l}{\textit{Proprietary (zero-shot generation)}} \\
GPT-5.4           & 51.0 & 42.3 & 37.0 & --- & 0 & .364 & .432 & .441 \\
GPT-5.4-mini      & 60.0 & 46.3 & 37.0 & --- & 0 & .360 & .424 & .436 \\
Opus 4.6          & \textbf{62.0} & \textbf{46.3} & \textbf{39.0} & --- & 0 & .379 & .411 & .426 \\
Sonnet 4.6        & 59.0 & 41.7 & 38.2 & --- & 0 & \textbf{.396} & \textbf{.437} & \textbf{.440} \\
\bottomrule
\end{tabular}
\par\vspace{0.3em}
{\footnotesize $^\dagger$FT/Vanilla: counted at 10-turn generation; proprietary: per-turn (all 0). ROUGE-L not computed for proprietary (different prompt format).}
\caption{Automatic evaluation of generated utterances (100 samples per generation length). Speaker accuracy = fraction of correctly predicted speakers averaged across all generated turns. Semantic similarity = cosine similarity of sentence embeddings (\texttt{all-MiniLM-L6-v2}) between concatenated generated and ground-truth turns. \textbf{Bold} = best per column.}
\label{tab:gen-auto}
\end{table*}

\paragraph{Surface post-processing.}
Manual inspection revealed that FT outputs, while contextually grounded, suffered from surface-level formatting issues absent in vanilla outputs, such as missing sentence-initial capitalization, missing punctuation, lowercase ``i'', and truncated sentence endings.
To ensure that evaluators judged content rather than formatting, we applied a two-pass surface polish to all FT utterances before evaluation.
Pass~1 used GPT-5.4-mini to correct capitalization and basic punctuation (1,390/1,440 utterances modified).
Pass~2 applied rule-based truncation fixes (278 trailing periods removed from incomplete sentences) followed by GPT-5.4-mini comma insertion (705 commas added).
All changes were verified as surface-only. No words were added, removed, or reordered.
Despite this polish, automated LLM judges still preferred Vanilla outputs (83--87\% across prompt variants), and the subsequent human evaluation confirmed the same pattern, indicating that the preference gap is not driven by surface formatting.

\paragraph{Quality filtering.}
We applied a 4-stage pipeline to select evaluation samples: (1)~GPT-based defect filtering (5 defect categories, 3-run majority vote, 2,700 evaluations $\rightarrow$ 171 pass), (2)~manual audit (3 removed), (3)~subtle quality filtering (unnatural enumeration), and (4)~uniform sampling (40 per generation length).
Five-turn samples required regeneration with improved parameters (temperature 0.7$\rightarrow$0.5, repetition penalty 1.2$\rightarrow$1.3, best-of-3 selection).
The final set comprises 120 samples $\times$ 3 pairs = 360 comparisons + 20 attention checks.

\paragraph{Human evaluation setup.}
We recruited evaluators through Prolific, requiring English fluency and prior experience with collaborative teamwork.
The evaluation interface presented pairs of AI-generated meeting continuations (A and B, randomly assigned) alongside the original conversation context.
Evaluators rated each pair on five dimensions: (1)~overall preference (A much better / A slightly better / Tie / B slightly better / B much better), (2)~confidence (1--5), and for each side separately, (3)~naturalness (1--5), (4)~coherence (1--5), and (5)~speaker consistency (1--5).

Quality control used two mechanisms: attention checks (3 per evaluator, requiring identification of a clearly superior continuation) and instructed-response checks (requiring a specific Likert value).
Evaluators failing $\geq$2 checks of the same type were terminated early.
Of 123~total participants, 66~passed quality thresholds, yielding 1,142~quality-controlled pairwise judgments across 360~comparison items (3~pairs $\times$ 120~samples).

\begin{figure}[ht]
    \centering
    \includegraphics[width=\linewidth]{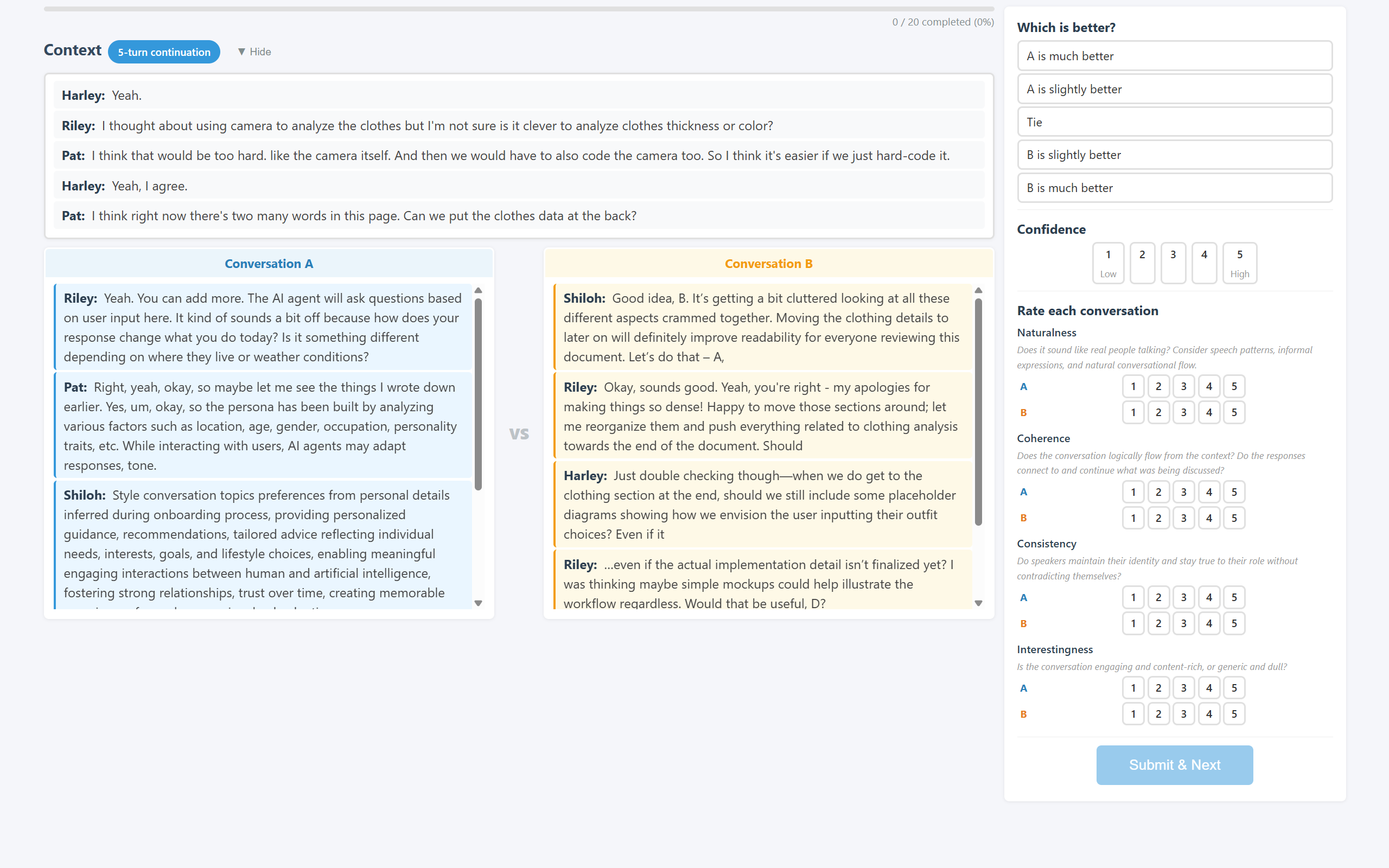}
    \caption{Human evaluation interface on Prolific. Evaluators view the original conversation context (top), two AI-generated continuations A and B (bottom left/right, randomly assigned), and rate overall preference, confidence, naturalness, coherence, speaker consistency, and interestingness on the right panel.}
    \label{fig:human-eval-interface}
\end{figure}

\subsection{LLM-as-Judge Evaluation}
\label{appendix:llm-judge}
Table~\ref{tab:llm-judge} compares two LLM judges.
GPT-5.4 strongly aligns with human preferences (Vanilla preferred in 66.7--72.5\% of comparisons involving Vanilla).
Opus~4.6 diverges: it prefers FT-Plain over Vanilla overall (54.2\% vs.\ 45.0\%) and FT-Reason over Vanilla (62.5\% vs.\ 37.5\%), particularly at shorter generation lengths.
At 1-turn, Opus prefers FT-Plain 72.5\% of the time, but this reverses at 5-turn where Vanilla is preferred 67.5\%.
This suggests that Opus is more sensitive to structural coherence (where FT excels) whereas GPT prioritizes surface fluency (where Vanilla excels), highlighting that LLM-based evaluation of multi-party dialogue generation remains model-dependent.

\begin{table}[ht]
\centering
\small
\begin{tabular}{lcc}
\toprule
& \multicolumn{2}{c}{\textbf{Win rate (\%)}} \\
\cmidrule(lr){2-3}
\textbf{Model} & \textbf{GPT-5.4} & \textbf{Opus 4.6} \\
\midrule
\multicolumn{3}{l}{\textit{FT-Plain vs.\ Vanilla}} \\
\quad FT-Plain & 27.5 & 54.2 \\
\quad Vanilla & 72.5 & 45.0 \\
\addlinespace
\multicolumn{3}{l}{\textit{FT-Reason vs.\ Vanilla}} \\
\quad FT-Reason & 33.3 & 62.5 \\
\quad Vanilla & 66.7 & 37.5 \\
\addlinespace
\multicolumn{3}{l}{\textit{FT-Plain vs.\ FT-Reason}} \\
\quad FT-Plain & 40.0 & 29.4 \\
\quad FT-Reason & 60.0 & 51.3 \\
\bottomrule
\end{tabular}
\par\vspace{0.3em}
{\footnotesize GPT-5.4 aligns with human evaluators (Vanilla preferred). Opus~4.6 diverges, preferring FT at shorter horizons.}
\caption{LLM-as-judge pairwise evaluation (360 comparisons per judge). Win rate = \% of comparisons in which the judge preferred that model. GPT-5.4 reports no ties; for Opus~4.6, the remainder within each comparison are ties.}
\label{tab:llm-judge}
\end{table}
\paragraph{Multi-turn fine-tuning failure.}
We attempted direct 5-turn fine-tuning (6,397 non-overlapping training samples), but 90\% of generated outputs failed to parse correctly.
We reverted to 1-turn fine-tuning with autoregressive multi-turn generation, which proved more reliable despite compounding errors over turns.

\section{Training and Inference Configuration}
\label{appendix:training-config}

Table~\ref{tab:training-config} summarizes the training hyperparameters for all fine-tuning experiments.
All models use LoRA adapters with completion-only loss (prompt tokens masked).
Inference for open-source models uses single-token logit scoring, where the prompt is fed through the model, and the candidate with the highest log-probability at the prediction position is selected.
For proprietary models, we use greedy generative decoding (temperature~0, max tokens~50) via the respective APIs.

\begin{table}[ht]
\centering
\small
\begin{tabular}{lcc}
\toprule
\textbf{Parameter} & \textbf{Gemma-3-12B} & \textbf{Llama-3.1-8B} \\
\midrule
LoRA rank ($r$) & 16 & 16 \\
LoRA alpha ($\alpha$) & 32 & 32 \\
LoRA dropout & 0 & 0 \\
LoRA targets & \multicolumn{2}{c}{q, k, v, o, gate, up, down\_proj} \\
Trainable params & $\sim$0.56\% & $\sim$0.56\% \\
Epochs & 3 (2 for Mixed) & 2 \\
Learning rate & 1e-4 & 2e-4 \\
LR scheduler & cosine & cosine \\
Warmup ratio & 0.1 & 0.1 \\
Batch size & 2 & 2 \\
Gradient accum. & 8 & 8 \\
Effective batch & 16 & 16 \\
Max seq length & 4,096 & 4,096 \\
Precision & bf16 & bf16 \\
Optimizer & AdamW & AdamW \\
Hardware & \multicolumn{2}{c}{2$\times$ NVIDIA RTX A6000 (48GB)} \\
Wall time (Exp~1) & $\sim$18h & $\sim$8.5h \\
\bottomrule
\end{tabular}
\caption{Training configuration for all fine-tuning experiments. All use LoRA with no quantization (full bf16).}
\label{tab:training-config}
\end{table}

\section{Prompt Templates}
\label{appendix:prompts}

We show the exact prompt format used for each task.
Both open-source and proprietary models receive identical system and user messages.

\paragraph{Speaker prediction (SU format).}
\begin{quote}
\small
\textbf{System:} \textit{You are an expert at predicting conversational dynamics in team meetings. Given a dialogue history, predict which team member will speak next. Answer with only the speaker ID (e.g., PERSON\_A).}

\textbf{User:} \textit{Below is a dialogue history from a team meeting. Predict which speaker speaks next.}

\textit{Dialogue:} \\
\textit{[Turn 1] PERSON\_A: Hello. I'm Jordan.} \\
\textit{[Turn 2] PERSON\_B: I'm Alex.} \\
\textit{[Turn 3] PERSON\_C: I'm Drew.} \\
\textit{[Turn 4] PERSON\_D: I'm Reese.} \\
\textit{[Turn 5] PERSON\_A: It's being recorded, so just in case, I'll turn this one on too.}

\textit{Participants in this meeting: PERSON\_C, PERSON\_B, PERSON\_A, PERSON\_D}

\textit{Who speaks next? Answer with only the speaker ID.}

\textbf{Expected output:} \textit{PERSON\_B}
\end{quote}

\paragraph{Intention prediction (SU format).}
\begin{quote}
\small
\textbf{System:} \textit{You are an expert at predicting utterance types in team meeting conversations. Given a dialogue history, predict the utterance type of the next turn.}

\textit{[14 category definitions provided, e.g.:}\\
\textit{- Active listening: Short utterances showing attention or agreement}\\
\textit{- Giving Information: Providing objective facts or asking factual questions}\\
\textit{...]}

\textit{Answer with only the utterance type name.}

\textbf{User:} \textit{[Same dialogue format as speaker prediction, with utterance types annotated per turn, e.g.:}\\
\textit{[Turn 1] PERSON\_A [Social / Humor]: Hello. I'm Jordan.]}

\textbf{Expected output:} \textit{Active listening}
\end{quote}

\paragraph{Generation (FT-Plain, plain format).}
The model receives the 5-turn context and generates the next utterance in ``\texttt{SPEAKER: utterance}'' format.
For FT-Reason (social-cue reasoning), the model first generates a structured chain: ``\texttt{Next speaker: X / Role: Y / Intention: Z / Utterance: text}''.
For Vanilla and proprietary models, the same context is provided, with instructions to continue the conversation naturally.

\end{document}